\documentclass[11pt]{article}
\usepackage[margin=1in]{geometry}
\usepackage{graphicx}
\usepackage{booktabs}
\usepackage{amssymb}
\newcommand{\cmark}{\ensuremath{\checkmark}}
\newcommand{\xmark}{\ensuremath{\times}}
\usepackage{amsmath}
\usepackage{multirow}
\usepackage{xcolor}
\usepackage[hidelinks]{hyperref}
\hypersetup{
  pdftitle={Srijika: OpenType-Layout-Reusing Font Restyling for Nine Indic Scripts},
  pdfauthor={Anil Pai},
  pdfsubject={Indic font restyling, glyph diffusion, and OpenType font assembly},
  pdfkeywords={font generation, font restyling, Indic scripts, OpenType, glyph diffusion, typography}
}
\usepackage{rotating}
\usepackage{tikz}
\usetikzlibrary{arrows.meta, positioning}

\newcommand{\srijika}{\textsc{Srijika}}
\newcommand{\lipika}{\textsc{Lipika}}

\title{\srijika: OpenType-Layout-Reusing Font Restyling\\
       for Nine Indic Scripts\\[0.4em]
       \large Retrieval-Grounded Glyph Diffusion, Template Replacement,\\
       \large and a Negative-Results Catalogue}

\author{Anil Pai\\
Loopdesk Technologies LLP\\
\texttt{anil@loopdesk.ai}}
\date{}

\begin{document}
\maketitle

\begin{abstract}
We present \srijika{}, a system that produces installable OpenType fonts
for nine Brahmic scripts --- Devanagari, Tamil, Bengali, Telugu, Kannada,
Malayalam, Gujarati, Gurmukhi, and Odia --- by restyling the glyph
outlines of shaping-complete template fonts rather than generating fonts
from scratch. An Indic font is a shaping-engine contract: hundreds to
thousands of conjuncts, half forms, and matra variants reached through
OpenType substitution must be present and mutually consistent, and raster
glyph generators stop short of this artifact. \emph{Template glyph
replacement} reuses the template's cmap and GSUB closure unchanged, and
its GPOS data under a documented metric policy, so every output is a font
by construction. The system ships 66 TTFs (57 curated presets and nine
open-vocabulary showcase fonts); all pass the OpenType Sanitizer,
HarfBuzz and CoreText reproduce the template's glyph-ID sequences on
conjunct-heavy probes for every font, and a full-closure audit (80{,}915
glyphs, 54{,}812 anchors) quantifies every metric delta. Style is
selected by natural language through \lipika{}, a retrieval index over
$\sim$650 open-license families that grounds a prompt in a concrete donor
face; a reference-conditioned latent diffusion model redraws each
template glyph in the donor's style; and a content gate, harmonization
pass, and shaped-cluster verify-and-repair loop revert any failing glyph
to its template outline. One warm-start graft recipe adds a script in
$\sim$20--48k steps; per-script routing sidesteps the cross-script
dilution grafting inflicts.

We evaluate against no-learning baselines only. On
diffusion-training-family-held-out SSIM
gates, unrestyled template copy beats generation on 50 of 56 faces; style
movement is measurable only with an internal, same-model embedding whose
training corpus includes the held-out families (median gap closure 0.05;
one OOD family, Alkatra, reaches 0.85--0.94 while seven others barely
move). A learned baseline, an independent style metric, and a human study
are outside this report's scope. Our contributions are the formulation,
the layout-reusing pipeline and its audit, a frozen nine-script
ID/OOD benchmark, and a negative-results catalogue: six unsuccessful
conditioning/objective levers, a forensic study of train-time objective
contamination, and a data-hull analysis of when reference-guided
restyling fails.
\end{abstract}

\section{Introduction}
More than 800 million people read Brahmic scripts (India's 2011 Census
alone records over a billion mother-tongue speakers of scheduled
languages written in them~\cite{census2011lang}), yet their typographic
ecosystems are a fraction of Latin's: Google Fonts hosts over 1,800
Latin families but, for a script like Odia, six families in
total~\cite{googlefonts2026}. The gap is structural. A
text-ready Indic font is not a drawing exercise but a shaping-engine
contract: hundreds to thousands of conjuncts, half forms, and matra
variants reachable through OpenType substitution must all be present and
mutually consistent, which multiplies the cost of every new design.
Generative models are the obvious lever, but the prevailing formulation
--- generate glyph \emph{images} --- stops short of the artifact anyone
can use. An image of type is not a font.

We present \srijika{}, a system that takes a natural-language style
description and emits an installable TTF for any of nine Brahmic
scripts, retaining the template's substitution closure and reusing its
positioning structure. Our central design decision is to \emph{never generate a
font from scratch}: we restyle the complete glyph closure of a
professionally-engineered open-license template font, tile by tile, with
a reference-conditioned latent diffusion model, then rebuild outlines
in place so the template's GSUB machinery is reused unchanged and its
GPOS structure is reused subject to documented metric and
anchor-coordinate transformations. Finite HarfBuzz and CoreText probes
verify glyph-ID parity with the template; visual positioning quality and
exhaustive GPOS behavior remain outside the present audit
(Sec.~\ref{sec:results}). Style
comes from \lipika{}, a retrieval index over $\sim$650 open-license
families whose embeddings ground open-vocabulary prompts in concrete
reference glyphs. A defect-tolerant serving pipeline --- content gate with
a retry ladder, style harmonization, coherence capping, and shaped-cluster
verification with per-glyph repair --- converts a model that is
\emph{usually} right into a font in which every glyph either passes
verification or reverts to the template outline.

Our contributions:
\begin{enumerate}
\item \textbf{Template-layout-reusing restyling}: the template's cmap
and GSUB substitution closure are retained unchanged; GPOS structure is
reused subject to documented metric and anchor-coordinate
transformations (Sec.~\ref{sec:system}). Layout is reused, not learned,
so restyling quality and layout logic decouple.
\item \textbf{Retrieval-grounded styling} (\lipika{}): per-(family,
script) CLIP-style embeddings turn free-text prompts into reference
tiles, sidestepping the absence of Indic style-attribute datasets
(Sec.~\ref{sec:system}).
\item \textbf{A graft recipe that scales one backbone to nine scripts}:
resume from a shipped checkpoint with uniform pack concatenation, plus a
\emph{dilution} observation --- under sequential grafting, previously
shipped scripts degrade relative to the donor, and re-weighted sampling
relocates rather than removes the loss (zero-sum in our runs).
Per-script routing, serving each script from the checkpoint that first
passed its gate, is our engineering response: it avoids shipped-script
regressions trivially, because shipped checkpoints are never updated ---
an operational choice, not a finding (Sec.~\ref{sec:training}).
\item \textbf{An evaluation protocol for generated Indic fonts}:
family-held-out zero-shot gates with genre-matched splits, script bars,
content probes, a shaped-cluster critic, and a frozen two-strata
(ID/style-OOD) benchmark with a diffusion-training-disjoint controlled
checkpoint (benchmark families remain present in \lipika{}'s encoder and
retrieval training corpus),
whose headline finding is that OOD style transfer is strongly
family-specific (Sec.~\ref{sec:eval}, Table~\ref{tab:bench}).
\item \textbf{A negative-results catalogue}: six tested-and-unsuccessful
lever families,
a forensic study of train-time objective contamination, and a data-hull
boundary --- failure modes characterized under our architecture, critic,
and training regime, which we expect (but do not show) to recur in similar
reference-guided glyph generators (Sec.~\ref{sec:negative}).
\end{enumerate}

\section{Related Work}
\textbf{Few-shot font generation.} Component- and style-disentangling
generators (FUNIT~\cite{funit2019}, MX-Font~\cite{mxfont2021},
DG-Font~\cite{dgfont2021}, CF-Font~\cite{cffont2023}) and diffusion approaches
(FontDiffuser~\cite{fontdiffuser2024}) synthesize unseen glyphs from a few
references, overwhelmingly for Latin and CJK. These methods emit rasters
and leave font assembly --- and for Brahmic scripts, the much harder
shaping problem --- out of scope. We adopt their reference-conditioning
insight but change the deliverable: our unit of generation is a
\emph{tile in a template's glyph closure}, so the output is an installable
font by construction.
\textbf{Direct vector-font synthesis.} A parallel line generates outlines
natively in vector space: VecFontSDF~\cite{vecfontsdf2023} reconstructs
quadratic outlines from signed distance fields,
DualVector~\cite{dualvector2023} learns dual-part B\'ezier representations
without vector supervision, VecFusion~\cite{vecfusion2024} applies a
raster-then-vector diffusion cascade, and VecGlypher~\cite{vecglypher2026}
emits glyph outlines with a language model. These systems generate
vectors natively and avoid a tracing stage, but target isolated glyphs ---
predominantly Latin/CJK --- and none addresses OpenType layout closure,
which for Brahmic scripts can exceed a thousand interdependent glyphs per
face (Sec.~\ref{sec:setting}). We chose raster diffusion plus tracing
because it composes with any template's closure out of the box; swapping
in a native vector generator per tile is compatible with our pipeline and
an attractive upgrade path.
\textbf{Diffusion conditioning.} Our generator is a latent diffusion
model~\cite{ldm2022} with reference cross-attention, related in spirit to
adapter-style conditioning~\cite{t2iadapter2023,controlnet2023}. Our
forensics run against the intuition that stronger decode-side content
injection is the lever for content fidelity: all six of our unsuccessful
levers attacked decode-side fidelity, and the evidence in our tested
setting points to training data coverage and objective hygiene as the
binding constraints
(Sec.~\ref{sec:negative}). We have not run a learned few-shot baseline
(e.g., FontDiffuser) on our pair packs, so this reading is scoped to our
architecture and is not yet a comparative claim
(Sec.~\ref{sec:limitations}).
\textbf{Font retrieval and attributes.} Attribute-based font selection
\cite{odonovan2014} and CLIP-based font embeddings
(FontCLIP~\cite{fontclip2024}) are trained on Latin attribute data (with
reported zero-shot generalization to other scripts); no purpose-built
Indic resource
existed. \lipika{} fills this gap with per-(family, script) embeddings
over an Indic corpus: a CLIP-LoRA tower for donor retrieval, and a
family-recognizer backbone used as the generator's style encoder and as
the critic's style metric.
\textbf{Indic text shaping.} Correct rendering of Brahmic scripts is
codified in Unicode~\cite{unicode16} and implemented by HarfBuzz~\cite{harfbuzz}. We rely
on it in two ways: templates are chosen from fonts whose shaping is
professionally validated, and our cluster critic re-shapes generated fonts
through HarfBuzz to check that shaped clusters remain visually plausible
after restyling.

\section{Problem Setting: Nine Brahmic Scripts}
\label{sec:setting}
Brahmic scripts multiply the difficulty of font generation along three
axes. \textbf{Closure size and shaping:} a text-ready font is not a set of
codepoint glyphs but the \emph{layout closure} of its script blocks ---
conjunct ligatures, half forms, matra variants, and contextual substitutes
reachable through GSUB. Closures vary an order of magnitude across scripts
(Mukta Mahee's Gurmukhi closure is 210 glyphs; Baloo Da~2's Bengali
closure is 1264), and any generated font must keep this machinery intact
or clusters visibly break. \textbf{Data scarcity with genre skew:} the
open-license corpus we could assemble spans $\sim$650 families across
nine scripts, but per-script family counts range from 250 (Devanagari)
down to 7 (Odia), and display/decorative genres are nearly absent outside
the largest scripts --- exactly the styles users ask for. \textbf{Visual
regime:} matras, stacked conjuncts, and headstrokes (\'sirorekh\=a) make
per-glyph plausibility insufficient; errors surface only when shaped text
is read as clusters, which motivates our cluster-level verification.

The corpus is drawn from Google Fonts (OFL/Apache), SMC, and community
foundries; every training font, template, and donor is open-licensed, and
generated fonts carry OFL-compliant renames crediting their template.

\section{The \srijika{} System}
\label{sec:system}
\subsection{Template glyph replacement}
Given a natural-language style description, \srijika{} does not synthesize
a font from nothing. It \emph{restyles} a curated, shaping-complete
\emph{template} font: for each glyph outline in the template's script
closure, a generative model redraws the glyph in the target style, and the
new outline is traced back into the template's \texttt{glyf} table. The
cmap and GSUB lookup data of the template are retained unchanged; GPOS
lookup structure and feature routing are retained, with anchors belonging
to a translated outline moved by the same vector and pair-positioning
values retained (the metric policy below). Because glyph selection is
driven by cmap and GSUB, a shaping engine makes the same
glyph-selection decisions for the output font as for the template
(Sec.~\ref{sec:eval} verifies this empirically); visual quality of the
restyled outlines is a separate question, handled by verification. This
converts
an open-ended font-engineering problem into a per-glyph image translation
problem with a structural floor: a rejected glyph reverts to its
professionally drawn template outline, while accepted generated outlines
remain subject to the limits of the internal critic
(Sec.~\ref{sec:results}).

Formally, let $T$ be the template with script glyph closure
$G(T)=\{g_1,\dots,g_n\}$ (computed by subsetting with layout closure over
the script's Unicode blocks), and let $R=\{r_1,\dots,r_6\}$ be six
reference tiles rendered from a \emph{donor} font that carries the target
style. A conditional generator $f_\theta(g_i, R)$ produces a restyled
raster tile for every $g_i$, which is thresholded, vectorized, and
re-inserted into the template's glyph slot with metrics derived from the
template's (plus a raster-space baseline alignment before vectorization,
below): the advance widens
--- never narrows --- when the restyled outline is fatter, and ink
falling left of the template's own sidebearing floor is shifted right
with the glyph's GPOS anchors translated by the same vector; pair-positioning
values are retained unchanged. This is an implementation policy that keeps
positioning data consistent with moved outlines --- it does not by itself
establish that every attachment remains visually appropriate for a
redesigned stroke or terminal (anchor-to-ink and collision auditing remain
future work, Sec.~\ref{sec:results}); output advances differ from
the template's only by these policy-defined widenings, whose measured
extrema are not typographic safety bounds
(Sec.~\ref{sec:results} measures the full distributions). One audited template per script sufficed to reach the fidelity
gates we report --- though it bounds layout-scale style transfer
(Sec.~\ref{sec:limitations}); we audit candidates by glyph
closure size and GSUB completeness (e.g.\ closures: Meera 1153 glyphs for
Malayalam, Baloo Da~2 1264 for Bengali, Mukta Mahee 210 for Gurmukhi ---
Gurmukhi closures are structurally small since the script has few
conjuncts).

\subsection{\lipika{}: grounding language in a donor font}
Open-vocabulary style control is delegated to retrieval. \lipika{}
comprises two image models trained on the same Indic font corpus. The
first is a \emph{recognizer}: a ConvNeXt-V2-tiny trained with a
sub-center ArcFace head to identify font families from rendered glyphs;
its frozen penultimate features serve as the generator's style encoder
(Sec.~\ref{sec:diffusion}), as the serving critic's style judge, and as
the embedding behind the closure metric of Sec.~\ref{sec:results} --- a
fact that makes that metric a same-model measurement, as discussed there.
The second grounds language: a
CLIP-L text/image tower pair with a LoRA-adapted~\cite{lora2022} image tower fine-tuned on
rendered specimens of the corpus, indexed with one L2-normalized
embedding per (family, script) face. A description (``warm rounded
friendly'') retrieves a ranked list of donor faces, optionally filtered by
script; the top face supplies the six style-reference tiles. The tower is
trained with a fused attribute-supervision stack: deterministic labels from
foundry metadata, measured stroke-contrast (EDT ridge widths on rendered
specimens, clamping VLM claims), and two VLM~\cite{qwen25vl} annotation passes over a
48-attribute vocabulary; the shipped checkpoint reaches a mean Spearman of
0.50 between attribute-text similarity and gold labels on held-out
families. Retrieval
grounding has two properties generation-from-text lacks: the style is
always \emph{realizable} (it exists as a font), and licensing is auditable
per donor.

\subsection{Reference-conditioned glyph diffusion}
\label{sec:diffusion}
The generator is a latent-diffusion UNet (134M parameters: 132.9M UNet +
1.2M style projection) over $128\times128$ glyph tiles.
The template glyph raster enters as a content channel at
\texttt{conv\_in}; the style is injected by cross-attention over pooled
embeddings of the six reference tiles, computed by the frozen \lipika{}
recognizer backbone (Sec.~\ref{sec:system}) --- only the 1.2M-parameter
projection onto cross-attention tokens is trained; classifier-free
guidance~\cite{cfg2022} is enabled
by token dropout at train time. Sampling uses a guidance scale of 3--4.5
and 50 steps. The same backbone, trained once and grafted per script
(Sec.~\ref{sec:training}), serves all nine scripts through per-script
routing.

\subsection{Defect-tolerant serving pipeline}
Raw per-glyph generation is not shippable; \srijika{} wraps it in four
verification-and-repair stages (Fig.~\ref{fig:pipeline}), each with a
template-fallback exit:

\begin{figure}[t]
\centering
\begin{tikzpicture}[
  font=\small,
  box/.style={draw, rounded corners=2pt, align=center, minimum height=9mm,
              inner sep=5pt, fill=blue!4},
  stage/.style={draw, rounded corners=2pt, align=center, minimum height=9mm,
                inner sep=5pt, fill=orange!8},
  io/.style={draw, rounded corners=2pt, align=center, minimum height=9mm,
             inner sep=5pt, fill=green!7},
  arr/.style={-{Stealth[length=2.4mm]}, thick},
  fb/.style={-{Stealth[length=2mm]}, densely dashed, gray},
  node distance=4mm and 7mm]

\node[io] (prompt) {prompt};
\node[box, right=of prompt] (lipika) {\lipika{}\\retrieval};
\node[box, right=of lipika] (refs) {donor refs\\+ precheck};
\node[box, right=of refs] (diff) {diffusion\\restyle};
\node[stage, right=of diff] (gate) {\textbf{1} content\\gate + retries};
\node[stage, right=of gate] (harm) {\textbf{2} harmonize\\+ coherence cap};
\node[stage, below=9mm of harm] (base) {\textbf{3} baseline\\snap};
\node[stage, left=of base] (verify) {\textbf{4} cluster\\verify + repair};
\node[box, left=of verify] (asm) {outline fit\\+ assembly};
\node[io, left=of asm] (ttf) {installable\\TTF};

\draw[arr] (prompt) -- (lipika);
\draw[arr] (lipika) -- (refs);
\draw[arr] (refs) -- (diff);
\draw[arr] (diff) -- (gate);
\draw[arr] (gate) -- (harm);
\draw[arr] (harm) -- (base);
\draw[arr] (base) -- (verify);
\draw[arr] (verify) -- (asm);
\draw[arr] (asm) -- (ttf);

\node[draw, dashed, gray, rounded corners=2pt, minimum width=28mm,
      minimum height=8mm, below=8mm of verify, align=center, font=\scriptsize]
      (tmpl) {template outline (fallback)};
\draw[fb] (gate.south) to[bend right=18] (tmpl.east);
\draw[fb] (harm.east) to[bend left=40] (tmpl.south east);
\draw[fb] (verify.south) -- (tmpl.north);
\end{tikzpicture}
\caption{The \srijika{} serving pipeline. A prompt retrieves donor
reference glyphs via \lipika{}; the diffusion model restyles the
template's glyph closure tile-by-tile; four verification stages gate,
harmonize, snap, and shape-verify the result, each able to revert
individual glyphs to the template outline (dashed) --- so every glyph
in an output font either passed verification or is the template
outline, and the layout program is inherited.}
\label{fig:pipeline}
\end{figure}
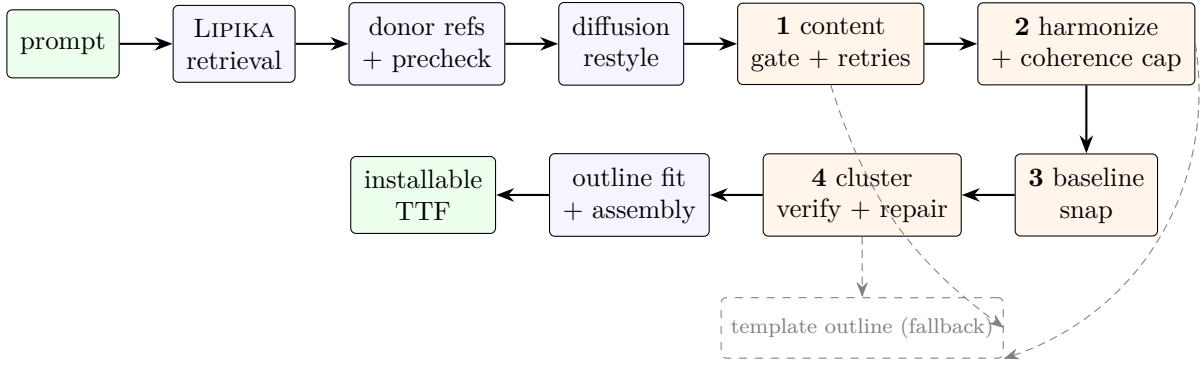

\textbf{(1) Content gate + retry ladder.} A retrieval-based
content-identity critic checks that each generated tile still reads as its
source glyph. Failing glyphs retry down a guidance ladder; unrecoverable
glyphs keep the template outline. (The critic is a \emph{serve-time
filter only}; adding it as a train-time loss destroys training ---
Sec.~\ref{sec:negative}.)

\textbf{(2) Weight harmonization with a bounded coherence cap.} Per-glyph
stroke-weight outliers (log ink-ratio deviations, MAD-scaled) are
regenerated with direction-aware guidance ladders (thin outliers retry at
higher guidance, thick at lower); a counter-fill guard (border flood-fill
hole area) catches blobbed counters that erosion statistics miss. A
blended-median coherence cap reverts to the template any glyph that remains
$>1.35\times$ above the fleet median weight --- but with \emph{bounded
authority}: if more than 25\% of styled tiles are flagged, the heaviness
is the style itself, and the cap stands down rather than delete the style
(\texttt{declined\_systemic}).

\textbf{(3) Baseline snap.} Per-glyph vertical jitter is cancelled in
raster space, before vectorization: row-profile cross-correlation against
the content raster measures each glyph's vertical drift, the font-wide
median drift is kept as style, and the residual jitter adjusts the trace
offset. Outlines therefore enter the template's coordinate system already
normalized --- no positioned outline is moved vertically afterwards, so
the template's anchor $y$-coordinates remain unchanged by baseline
normalization (the audit of Sec.~\ref{sec:results} confirms
$\Delta y \equiv 0$ across
all 54{,}812 anchors); whether unchanged coordinates remain visually
appropriate for redesigned ink is a separate question, evaluated only
through the shaped-cluster critic so far. Horizontal sidebearing shifts,
by contrast, happen
in vector space and translate anchors explicitly (Sec.~\ref{sec:system}).

\textbf{(4) Shaped-cluster verify and repair.} The assembled font is
shaped with HarfBuzz over the script's cluster inventory; clusters that
no longer read correctly (content critic on the shaped raster) have their
constituent glyphs greedily reverted to template outlines until the
cluster reads again. Fleet-wide this converges at cluster accuracy
0.981--1.000 with 7--110 reverted glyphs per font.

Finally, a \emph{donor precheck} measures the donor's bbox-normalized ink
density and warns when it exceeds the training hull
(Sec.~\ref{sec:hull}), and OFL-compliant renaming credits the template.

\subsection{Parametric post-axes}
Shipped fonts expose three outline-space parametric axes (weight, counter,
em-fill) applied directly to the TTF, giving users a small design space
around each generated style without re-sampling the model.

\section{Training: One Graft Recipe, Nine Scripts}
\label{sec:training}
\subsection{Pair packs and curated splits}
Training data is generated from the corpus itself: within each family,
(content, style) tile pairs are rendered where the content channel comes
from the script's template and the target comes from a corpus font, over
per-script shaping-aware unit inventories (aksharas, conjuncts, matra
combinations). Packs per script range from $\sim$8k to $\sim$300k+ pairs
across 7--250 families. A glyph-level auxiliary pack (single codepoints
rendered per font) hardens coverage; a script-sniffing bug taught us to
validate the per-script Unicode ranges first (glyph-o.k.\ rates of
62--87\% when correct; 2--3\% when silently falling back to Devanagari
codepoints).

\begin{figure}[t!]
\centering
\includegraphics[width=0.82\linewidth]{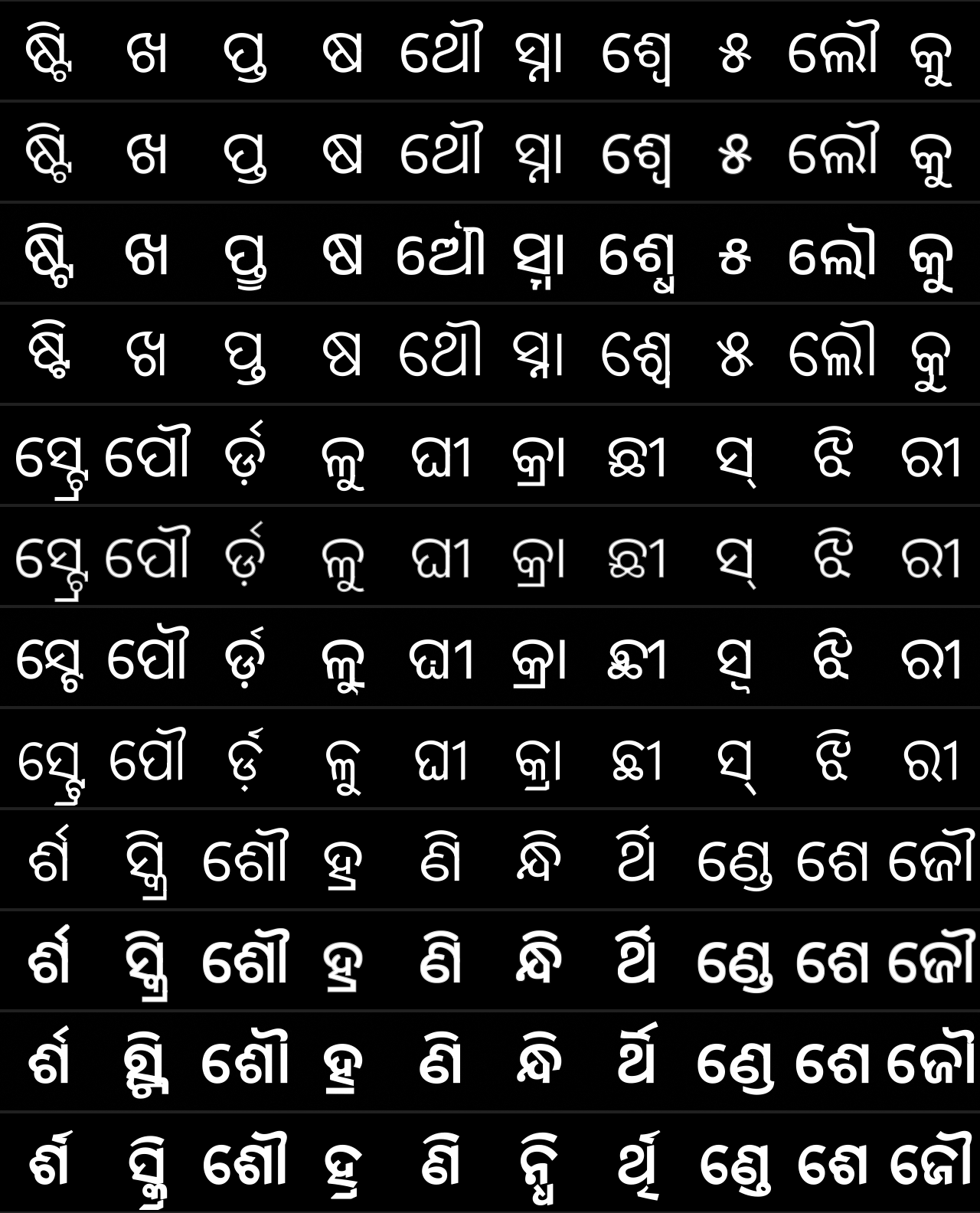}
\caption{Zero-shot reconstruction of held-out Odia faces (gen-v3.11b;
Lohit Oriya and two Noto Serif Oriya weights). Each block shows, top to
bottom: the template glyphs (model input content), the target-oracle
morphology reference (template dilated/eroded to the target's ink
coverage), generated tiles conditioned on six reference glyphs from the
unseen family, and the ground-truth target. Template and morphology are
nearly indistinguishable; the generated row moves stroke weight,
terminals, and contrast toward the target.}
\label{fig:odiarecon}
\end{figure}

Split curation matters more than volume. Our deployment policy ---
adopted after twice observing the same failure, and recorded as the
\emph{Modak rule} --- keeps decorative/display families in the training
split: in our runs they were unlearnable zero-shot and
depressed fresh-script baselines when placed in test. The Odia expansion
provides a two-run demonstration: with the decorative Alkatra family in
test, the script gated at 0.6541 SSIM while its reconstruction grids
looked structurally correct and a normal-genre validation family scored
0.720; re-splitting Alkatra into train and a serif family into test, then
retraining with the same recipe and initialization (both runs resume the
same v3.7 checkpoint; only split membership differs), yielded a 0.7622
zero-shot gate. The $+0.108$ delta is a \emph{benchmark-composition
difference} --- test distribution and training membership both changed ---
not a like-for-like model improvement. An instrumented re-run of both
conditions (Table~\ref{tab:alkatra}, now by held-out family) sharpens
the reading: for the held-out Alkatra family, generated \lipika{} gap
closure is 0.866 (seed-stable at 0.866/0.861/0.867), compared with 0.009
for the tile-space geometric transform and 0.008 for morphology; the
two-family OOD-condition macro is 0.450. This closure is comparable to,
not exceeding, the largest in-distribution family closure (0.98), and it
arrives while SSIM falls below even template copy --- qualitative
support (not controlled validation) that the Alkatra case exhibits
visibly larger local restyling and high internal \lipika{} closure,
despite lower SSIM --- a case the
in-distribution SSIM gate penalizes rather than rewards. We note the epistemic status
plainly: this is a post-hoc test-set change, applying a rule established
earlier on Devanagari rather than tuned per script, and we report both
numbers --- 0.6541 with the decorative family held out (in effect an
out-of-distribution challenge set) and 0.7622 genre-matched --- rather
than only the favorable one (Table~\ref{tab:alkatra}).
Operationally the Modak rule is a split policy; scientifically, excluding
difficult genres from test is not a satisfying evaluation design ---
decorative styles are precisely what users ask for. This post-hoc Odia
case motivates a controlled two-strata protocol with genre-matched and
style-OOD families reported side by side. Table~\ref{tab:alkatra}
records the historical Odia study; \texttt{benchmark\_v1} is the
controlled nine-script realization: a versioned
benchmark manifest (one decorative/display OOD family per script ---
Modak for Devanagari, Alkatra for Bengali and Odia, Akaya for Telugu and
Kannada, Coiny, Chilanka, Shrikhand, Langar --- alongside the existing
genre-matched ID families, with templates, reference-selection rule,
seeds, metrics, and SHA-256 hashes) ships in the supplement (the arXiv
ancillary files), and
Sec.~\ref{sec:results} reports the single-model two-strata evaluation on
it (Table~\ref{tab:bench}).
Fig.~\ref{fig:odiarecon} shows the
resulting held-out reconstructions.

\begin{table}[h]
\centering
\small
\caption{Odia split-composition study, by held-out family (the
independent unit): two training runs with the same
recipe and initialization (gen-v3.11 / gen-v3.11b, both resumed from
v3.7), differing only in which family is held out (and, symmetrically, in
training membership). All methods under the instrumented protocol of
Sec.~\ref{sec:results}: family-mean SSIM for generation, template copy,
target-oracle morphology, and the reference-only tile-space geometric
transform, plus family \lipika{} similarity-gap closures for generation,
geometry, and morphology; \emph{condition macro} rows average the two
family rows unweighted (per-face pooled SSIM, the gate statistic of
Table~\ref{tab:gates}, is 0.6541 OOD / 0.7622 genre-matched). For the
held-out Alkatra family, generated closure is 0.866 (0.866/0.861/0.867
across three diffusion seeds), versus 0.009 for geometry and 0.008 for
morphology; the two-family OOD-condition macro is 0.450. The decorative
hold-out is in effect an out-of-distribution challenge; the genre-matched
hold-out is the shipped gate. The re-split is post-hoc
(Sec.~\ref{sec:training}); both conditions are part of the record, the
delta is a benchmark-composition difference between two separately
trained runs --- the controlled single-model version of this comparison
is Table~\ref{tab:bench}, which replicates the effect. The OOD row here is \emph{qualitative
support}, not controlled validation: SSIM ranks the visibly style-moving
output below template-like structure while neither tested baseline
reproduces the movement (Fig.~\ref{fig:alkatra}).}
\label{tab:alkatra}
\setlength{\tabcolsep}{3.4pt}
\begin{tabular}{llcccccccc}
\toprule
& & & \multicolumn{4}{c}{SSIM} & \multicolumn{3}{c}{Fam.\ \lipika{} closure} \\
\cmidrule(lr){4-7}\cmidrule(lr){8-10}
Condition & Held-out family & Faces & Gen & Tpl & Morph & Geom &
Gen & Geo & Morph \\
\midrule
OOD (gen-v3.11) & Alkatra & 5 & 0.6377 & 0.6997 & 0.6922 & 0.7101 & 0.866 & 0.009 & 0.008 \\
 & Lohit Oriya & 1 & 0.7364 & 0.7619 & 0.7628 & 0.7325 & 0.034 & 0.020 & $-$0.003 \\
 & \emph{condition macro} & & 0.6870 & 0.7308 & 0.7275 & 0.7213 & 0.450 & 0.014 & 0.002 \\
\addlinespace[2pt]
Genre (gen-v3.11b) & Noto Serif Oriya & 5 & 0.7689 & 0.7917 & 0.7903 & 0.7608 & 0.025 & 0.006 & 0.007 \\
 & Lohit Oriya & 1 & 0.7290 & 0.7619 & 0.7628 & 0.7325 & 0.026 & 0.020 & $-$0.003 \\
 & \emph{condition macro} & & 0.7490 & 0.7768 & 0.7766 & 0.7466 & 0.026 & 0.013 & 0.002 \\
\bottomrule
\end{tabular}
\end{table}

\begin{figure}[t!]
\centering
\includegraphics[width=0.8\linewidth]{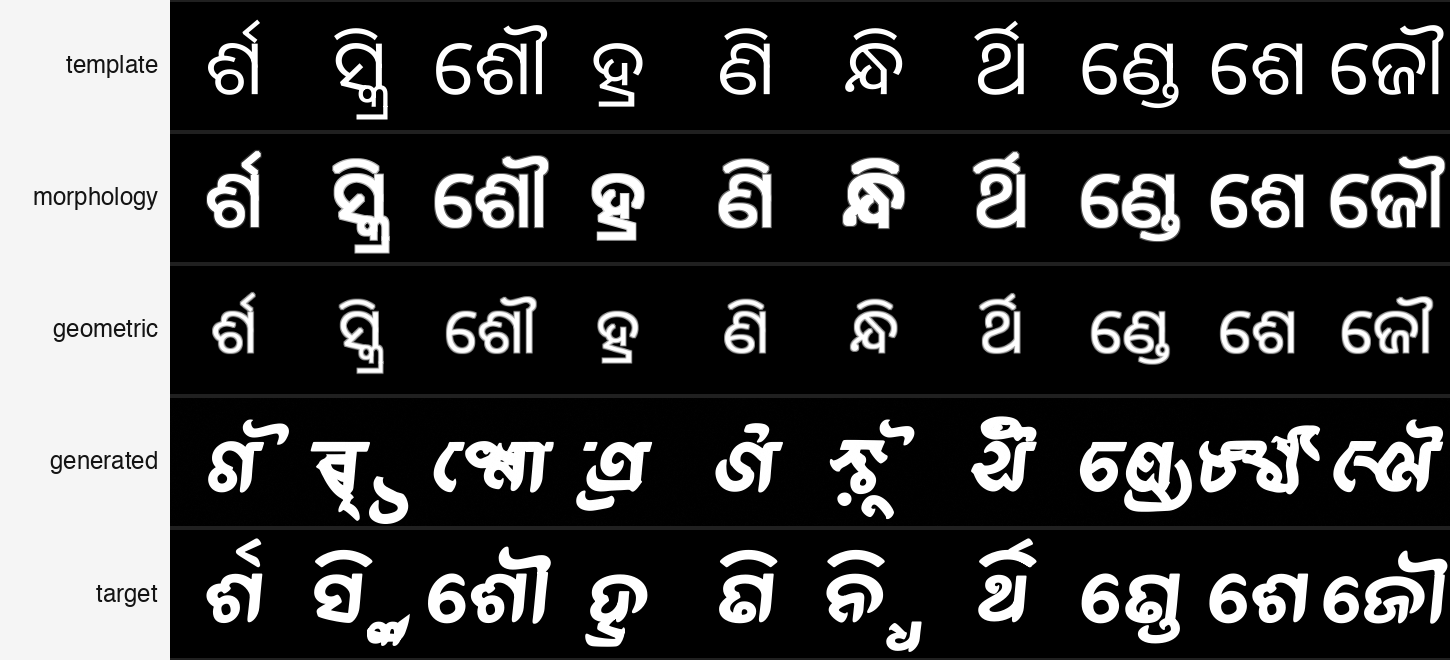}\\[2pt]
\includegraphics[width=0.5\linewidth]{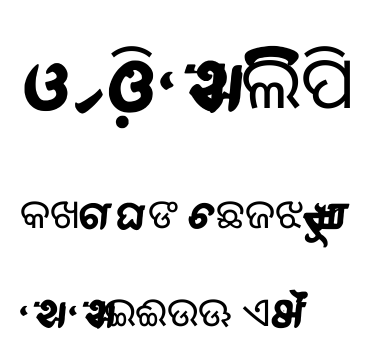}
\caption{The OOD condition, qualitatively. Top: the held-out Alkatra
wght700 face under gen-v3.11, rows labeled in-figure: template,
target-oracle morphology, reference-only tile-space geometric transform,
generated tiles, and the Alkatra ground truth. Morphology and
geometry barely move from the template --- consistent with their
near-zero \lipika{} closures in Table~\ref{tab:alkatra} --- while
generation visibly adopts the donor's heavy rounded style at a
structural cost that SSIM penalizes. Bottom: a HarfBuzz-shaped specimen
of a font assembled from the \emph{same held-out checkpoint and
conditioning} as the raster rows (gen-v3.11, seed 7, guidance 3,
template Noto Sans Oriya wght400, the eval protocol's six Alkatra-wght700
reference tiles): 210 of 562 closure glyphs passed the content gate and
are restyled, the remainder revert to template outlines --- the visible
weight mixing is the pipeline's honest OOD behavior, not a curated
production preset. The artifact passes \texttt{ots}; SHA-256
\texttt{03560c71c59f9614}\ldots{} in the supplement manifest.}
\label{fig:alkatra}
\end{figure}

\subsection{Warm-start grafting}
Every script after the first three is added by the same recipe: resume
the latest passing checkpoint (weights, optimizer, EMA), append the new
script's packs to the mix by uniform concatenation, and continue for a
20--48k-step window with a cosine LR restart (peak $\approx$
$2.3\times10^{-5}$, floor $0.05\times$). New-script quality tracks the
donor checkpoint's competence: Gurmukhi and Odia entered at 0.7861 and
0.7622 zero-shot SSIM --- above several first-generation script baselines
--- for $\sim$7 GPU-hours each on a single H100.

\subsection{Dilution under grafting and per-script routing}
Grafting has a cost: when the appended data is a large fraction of the
mix (+42\% for Malayalam+Gujarati), most previously shipped scripts
degrade \emph{relative to the donor} at both window lengths
(Table~\ref{tab:dilution}). The degradation is not monotone in window
length --- Tamil and Kannada partially recover between +20k and +30k,
Kannada to just above the donor --- so a longer window is not a
reliable fix,
and re-weighted sampling only relocates it --- zero-sum --- rather than
removing it (v3.12 column; controlled isolation in
Sec.~\ref{sec:negative}). We call this regularity \emph{dilution}; it is
an observation under one architecture, capacity, data order, and
schedule, not a claimed universal law. Our response is
\emph{per-script routing}:
each script is served by the checkpoint that first passed its gate
(v3.7 for the first five scripts; v3.8b for Malayalam/Gujarati; v3.11b
for Gurmukhi/Odia). Routing is an engineering choice, not a finding:
a shipped checkpoint is never updated, so it cannot regress, trivially.
It costs one resident model per graft generation, and it leaves the
scientific question --- a single model that adds scripts without
dilution --- open (Sec.~\ref{sec:limitations}).

\begin{table}[h]
\centering
\caption{Dilution under grafting. Grafting Malayalam+Gujarati (+42\% pairs,
uniform sampling) onto gen-v3.7 (@120k) degrades most previously shipped
scripts relative to the donor at both window lengths, while new scripts
improve. The degradation is not monotone in window length --- Tamil and
Kannada partially recover from +20k to +30k, Kannada to just above the
donor --- so a longer window is not a reliable fix.
Re-weighted sampling over the same
window (v3.12, shares restoring each old script's pre-graft proportion)
shifts mean SSIM by $\approx 0$ vs.\ v3.8b and relocates the damage to
the down-weighted scripts: in our runs dilution is zero-sum. Old-script
regressions
in \textbf{bold}.}
\label{tab:dilution}
\begin{tabular}{lcccc}
\toprule
 & v3.7 (donor) & v3.8 (+20k) & v3.8b (+30k) & v3.12 (+30k, wtd.) \\
\midrule
Devanagari & 0.7476 & \textbf{0.7458} & \textbf{0.7408} & \textbf{0.7408} \\
Tamil      & 0.7570 & \textbf{0.7493} & \textbf{0.7532} & \textbf{0.7498} \\
Bengali    & 0.7687 & \textbf{0.7558} & \textbf{0.7534} & \textbf{0.7587} \\
Telugu     & 0.7691 & 0.7824          & 0.7790 & 0.7828 \\
Kannada    & 0.6860 & \textbf{0.6714} & 0.6887 & \textbf{0.6803} \\
Malayalam  & ---    & 0.7346          & 0.7401 & \textbf{0.7298} \\
Gujarati   & ---    & 0.7022          & 0.7019 & 0.7024 \\
\bottomrule
\end{tabular}
\end{table}

\section{Evaluation Protocol}
\label{sec:eval}
\textbf{Zero-shot gates.} Each script holds out complete font families
(never individual weights) as a test split, genre-matched per the Modak
rule. At evaluation, the model restyles the script's template glyphs using
six reference tiles from each held-out family and scores SSIM between
generated and ground-truth tiles over the script's unit inventory at
guidance 3. The reference tiles are drawn from a disjoint subset of the
unit inventory and are never among the scored targets. Table~\ref{tab:gates}
reports means pooled over all generated tiles of the
held-out faces. Tiles within a face are correlated, and faces of one
family are too, so pooled statistics overstate the effective sample size
($n$ is \emph{families}: 2--4 per script; faces are 2--12);
Table~\ref{tab:gates} therefore also reports family counts and
family-level macro-averages, and Appendix~\ref{app:perface} reports every
held-out face individually (the design is balanced at 64 tiles per face,
so the pooled mean equals the face-level macro-average), alongside
template-copy and density-matched morphology references and a
reference-only geometric baseline scored on
the same targets. This gate isolates \emph{reference-conditioned
reconstruction under oracle donor references} --- the donor is a real
held-out font, which removes prompt retrieval from the loop --- so it
neither bounds nor measures end-to-end
prompt-to-font quality; the end-to-end critic below and designer review
provide only partial evidence there, and a controlled human evaluation
of prompt-to-font output remains future work.
\textbf{Script bars.} A script's \emph{bar} is the gate score of the first
model that shipped it; no later model may serve a script below its bar
(quality-over-cost rule). This turns every subsequent training run into a
regression test across all shipped scripts.
\textbf{Content probes.} Two calibrated Latin display donors (Lobster:
heavy-but-inside-hull; AlfaSlab: beyond-hull) probe content survival under
aggressive styling: initial wrong-glyph counts and template-fallback rates
under the serving gate's retry ladder.
\textbf{End-to-end critic.} Shipped fonts are scored by donor style match
(embedding cosine), legibility fraction, and shaped-cluster accuracy after
repair; specimens are additionally proofed visually.

\section{Results}
\label{sec:results}
\begin{table}[h]
\centering
\caption{Shipped zero-shot gate scores: an \emph{in-distribution,
genre-matched operational gate} (SSIM over held-out families, test split,
guidance 3), read against three no-learning references --- template copy
(unrestyled template glyphs scored on the same targets),
density-matched morphology (template glyphs dilated/eroded to each
target's ink coverage; a \emph{target-oracle} diagnostic, since the
generator never sees the scored target), and Geom the coherent
reference-only geometric baseline (Sec.~\ref{sec:results}) --- plus the per-script serving route.
\emph{Fam.}\ counts held-out families (the independent unit; faces of one
family are correlated), \emph{Faces} the held-out faces, and \emph{Fam.\
macro} the family-level macro-average SSIM. Bars inherited from
the first shipping generation (Devanagari 0.7445, Tamil 0.7551, Bengali
0.7687) are met or exceeded; for the remaining scripts the serving model
is the first to ship them, so bar $=$ gate. Per-face scores and ranges
are in Appendix~\ref{app:perface}; family-level results for all methods
and paired style scores are in Table~\ref{tab:family}.}
\label{tab:gates}
\small
\begin{tabular}{lrrcccccl}
\toprule
Script & Fam. & Faces & SSIM & Fam.\ macro & Tpl.\ copy & Morph & Geom & Serving model \\
\midrule
Devanagari & 3 & 12 & 0.7476 & 0.7371 & 0.7913 & 0.7919 & 0.7787 & gen-v3.7 \\
Tamil      & 2 &  3 & 0.7570 & 0.7674 & 0.7598 & 0.7628 & 0.6918 & gen-v3.7 \\
Bengali    & 2 &  4 & 0.7687 & 0.7694 & 0.7810 & 0.7781 & 0.7653 & gen-v3.7 \\
Telugu     & 4 &  8 & 0.7691 & 0.7542 & 0.7797 & 0.7972 & 0.8356 & gen-v3.7 \\
Kannada    & 2 & 11 & 0.6860 & 0.6980 & 0.7546 & 0.7545 & 0.7481 & gen-v3.7 \\
Malayalam  & 4 &  7 & 0.7401 & 0.7397 & 0.7816 & 0.7820 & 0.7991 & gen-v3.8b \\
Gujarati   & 2 &  2 & 0.7019 & 0.7019 & 0.7260 & 0.7199 & 0.7245 & gen-v3.8b \\
Gurmukhi   & 2 &  3 & 0.7861 & 0.7779 & 0.8041 & 0.8077 & 0.7736 & gen-v3.11b \\
Odia       & 2 &  6 & 0.7622 & 0.7490 & 0.7867 & 0.7858 & 0.7561 & gen-v3.11b \\
\bottomrule
\end{tabular}
\end{table}

\paragraph{Zero-shot gates.} Table~\ref{tab:gates} reports the shipped
matrix. Three regularities: (i) scores track \emph{test-split genre
composition} more than script complexity --- Gurmukhi, added last, posts
the highest score not because it is easy but because its held-out
faces are body-text designs inside the training hull, while Kannada's
test split includes a display cut; (ii) held-out sets are small (2--4
families per script) and tiles within a face are correlated, so alongside
pooled means we report family-level macro-averages (Table~\ref{tab:gates})
and full per-face scores with p25 tails (Appendix~\ref{app:perface});
family macro-averages move pooled means by at most $\pm 0.015$ (largest
where a single family contributes 10 of 11 faces, Kannada), and we
caution that with $n{=}2$--$4$ independent families per script these are
descriptive statistics, not estimates with useful confidence intervals
(sampling-seed reruns with two additional diffusion seeds move pooled
per-script means by at most $0.006$ SSIM, Table~\ref{tab:family});
(iii) under these small descriptive test sets, the grafted scripts'
(Malayalam through Odia) family-macro gate scores overlap the range of
the original five under the identical protocol.

\paragraph{Template-copy baseline and a target-oracle diagnostic.} The
do-nothing baseline scores the unrestyled template glyph against the same
held-out targets. It posts \emph{higher} pooled SSIM than the model on
every script (by 0.003--0.069) and on 50 of 56 held-out faces
(Appendix~\ref{app:perface}). A second no-learning reference,
density-matched morphology, dilates or erodes each template glyph
(greyscale min/max filtering, blending bracketing radii) until its ink
coverage exactly matches the target tile's, then scores the morphed tile
--- it isolates how much of the SSIM gap is pure ink-coverage matching.
It is a \emph{target-oracle diagnostic}, not a comparable
reference-conditioned baseline: it reads each scored target's own ink
coverage, which the generator never receives (its conditioning is six
disjoint reference tiles), and its per-tile morphs are independently
tuned, so they need not assemble into a coherent font.
It lands within $\pm 0.018$ of template copy on every script
(Table~\ref{tab:gates}) and above generated SSIM on 53 of 56 faces:
density matching alone neither improves much on the raw template nor
lets generation catch up on in-distribution targets. This is expected and diagnostic: most
held-out faces are body-text designs structurally close to the
professionally drawn template, and SSIM rewards that proximity more than
it rewards style movement. On the \lipika{} style metric, generated
tiles attain 0.14--0.67 pooled normalized similarity to the held-out
face's six-reference centroid, versus 0.06--0.17 for the unrestyled
template under the same normalization
(Appendix~\ref{app:perface} gives the exact formula, per-face
values, and a paired \lipika{} similarity-gap-closure variant). At family
level the paired comparison separates the two questions --- and gives a
sober magnitude. Generated tiles sit closer to the reference centroid
than the template on 18 of 23 held-out families, but the median
\lipika{} similarity-gap closure is only 0.047 (IQR 0.016--0.133): 8
families exceed 0.10, 3 exceed 0.25, 1 exceeds 0.5, and 5 are negative
(minimum $-0.101$). Generation improves the internal \lipika{} style
score on most held-out families, yet the median family closes under
5\% of the template-to-target similarity gap, and the large closures
concentrate in a few families (Karla Tamil Upright 0.40, Meera Inimai
0.98, RIT TN Joy 0.26); template copy retains higher SSIM on 20 of 23
(Table~\ref{tab:family}) --- the model moves style at some structural
cost, and typically by a small amount. These statistics are seed-stable:
across the three sampling seeds the median closure is
0.047/0.054/0.047, the counts above 0.10 and 0.25 are unchanged, and
exactly one near-zero family (Lohit Malayalam) flips sign. The
morphology reference's style cosine
tracks the template's within $0.08$ on every family, so this style
movement is not reproduced by the tested ink-coverage-matched morphology
diagnostic nor, as the next baseline shows, by coherent reference-derived
geometry. The third no-learning reference is exactly that test: a
\emph{coherent reference-only geometric baseline} estimates one global
parameter set --- ink-bbox $x$/$y$ scale (clamped to $[0.8,1.25]$) plus
one fractional morphological weight radius --- from the six reference
tiles alone, the same conditioning the generator receives, and applies it
to every template tile (full estimation specification in
Appendix~\ref{app:perface}; the implementation is not included with
this version and is scheduled for a later release --- see Code and Data
Availability). Structurally it is competitive: it exceeds
generated SSIM on 17 of 23 families and template copy on 10 of 23
(pooled per script 0.69--0.84, Table~\ref{tab:gates}). On style it is
not: its median \lipika{} similarity-gap closure is 0.006 (IQR
$-0.004$--$0.027$, maximum 0.218, none above 0.25) versus generation's
0.047 median and 3 families above 0.25, and generation out-closes it on
17 of 23 families (Table~\ref{tab:family}). The style movement diffusion
delivers is thus not explained by global scale-and-weight geometry
either --- though richer parametric families (contrast, slant,
per-stroke modulation) remain untested
(Sec.~\ref{sec:limitations}). Where a held-out face departs far from the template the SSIM
ordering flips (Meera Inimai: ${+}0.043$ over template copy, ${+}0.039$
over morphology, at style
ratio 0.98). We therefore read Table~\ref{tab:gates} through its
baselines: the gates are in-distribution regression tests, meaningful
for run-over-run comparison (bars, Table~\ref{tab:dilution}) --- not
evidence that generation out-reconstructs copying on typical body-text
targets, and not a measure of open-vocabulary output quality
(Sec.~\ref{sec:limitations}).

\paragraph{Controlled two-strata benchmark.} To remove the post-hoc
confound of the Odia study (Table~\ref{tab:alkatra}) we trained one
model from scratch on the frozen \texttt{benchmark\_v1} manifests
(gen-bench-v1: all nine packs uniformly concatenated, 168k steps, no
resume, no auxiliary losses) with every script's ID, OOD, and validation
families excluded simultaneously from diffusion pair-pack training, then
evaluated both strata under
identical weights, references, seeds, and baselines
(Table~\ref{tab:bench}). Scope notes, stated plainly: the OOD families
were \emph{retrospectively selected} (Alkatra's behavior was known
before the manifest was written) and \emph{subsequently frozen} ---
this is controlled reproduction, not a preregistered discovery test;
the seed columns vary \emph{inference sampling only}, and gen-bench-v1
is one training run (training-seed replication was not run); the 168k
step budget and final-checkpoint rule were fixed in the frozen manifest
before training, and the test strata were never inspected during
diffusion-model fitting or checkpoint selection; and gen-bench-v1 is a
\emph{controlled research checkpoint} for studying generalization to
families held out of diffusion training, not the routed production
model. Three findings. (i)~\emph{The Alkatra case reproduces under
control}: the same held-out multilingual Alkatra family produces high
closure in both its Bengali (0.935) and Odia (0.853) script renderings
--- correlated script-level observations of one style family, not
independent family replications --- seed-stable, with geometry and
morphology at ${\leq}0.011$,
while SSIM stays below template copy; the pattern of
Table~\ref{tab:alkatra} is not an artifact of comparing two separately
trained runs. (ii)~\emph{OOD style movement is family-specific, not
category-general}: the other seven unique OOD families (Modak, Coiny, two
Akayas, Chilanka, Shrikhand, Langar) close a median 0.040 (range
$-0.034$--$0.062$) --- the decorative stratum is not uniformly
transferable from six references, and Alkatra's heavy rounded style is
the outlier the model imitates well, a caution against generalizing any
single OOD case. (iii)~\emph{Freezing both strata did not collapse
in-distribution behavior}: the frozen model's 22 ID families post a
closure median of 0.118 with SSIM$-$template deltas (median $-0.023$)
matching the shipped models' band; direct ranking against the serving
chain is confounded by different training budgets and mixes, so we
report both without ranking. One disjointness caveat applies to all
closure numbers: the exclusions above cover \emph{diffusion training};
a family-membership audit finds all 30 benchmark families (and both
Alkatra script renderings) inside the 595-family corpus that trained
both \lipika{} image models --- the CLIP retrieval tower and index, and
the recognizer whose frozen features are the generator's style encoder
and the closure metric's embedding. The judge is thus a family-identity
classifier trained to cluster the very families it scores, and the
generator's conditioning pathway carries a learned representation of
them. Representational exposure through
\lipika{} is therefore not excluded, and closure remains an internal
metric; evaluating with a benchmark-disjoint \lipika{} retrain and an
independent style embedding is future work, as is any human evaluation.
All 31 family rows, three seeds each, are
in Table~\ref{tab:bench} and the supplement CSV.

\begin{figure}[t!]
\centering
\includegraphics[width=\linewidth]{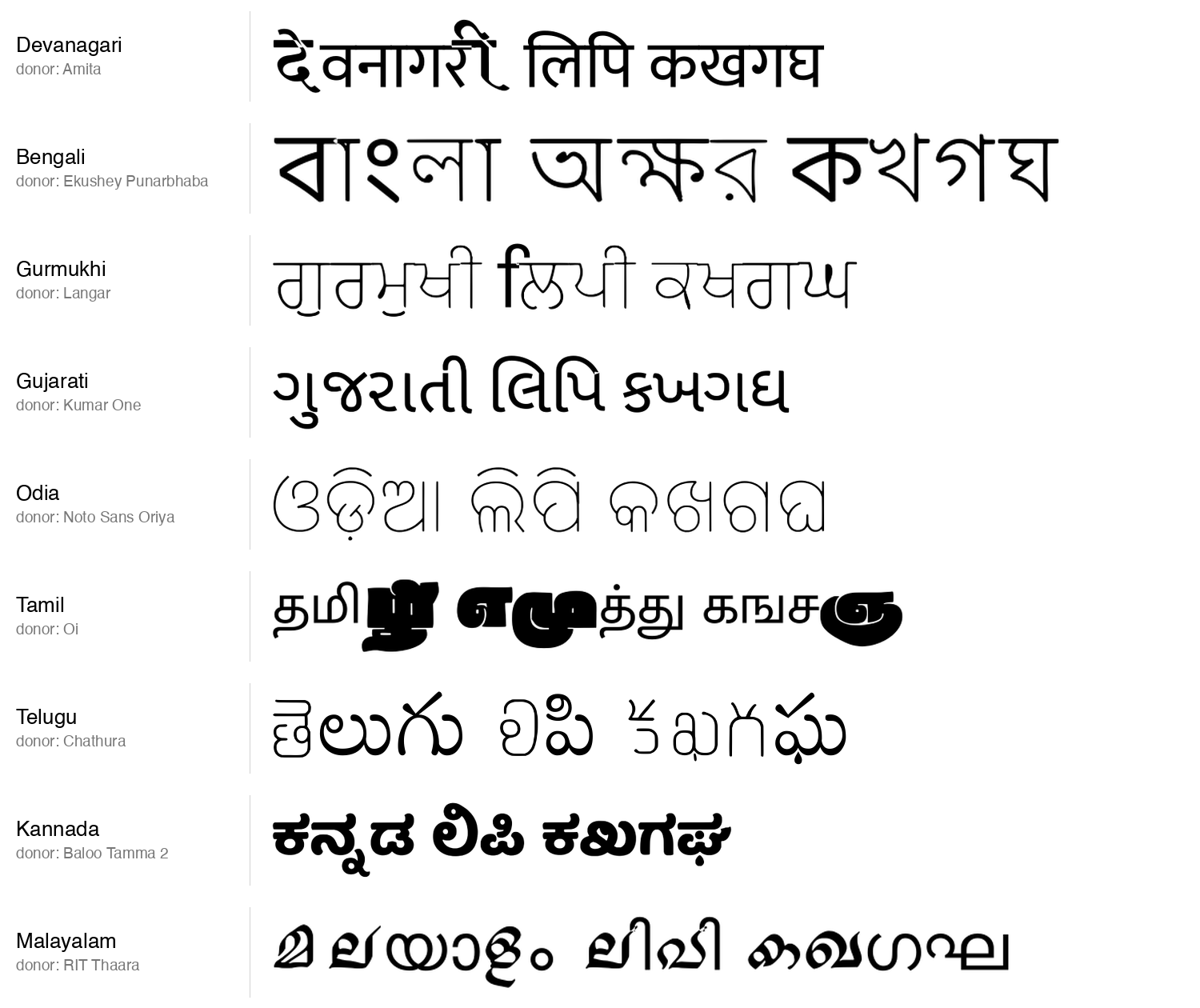}
\caption{One generated font per script from a single showcase run, each
rendered from its installable TTF with HarfBuzz shaping. Donors span
native-script families (Amita, Ekushey Punarbhaba, Kumar One, Baloo
Tamma~2, RIT Thaara) and system templates. The Tamil row (donor Oi, an
ultra-black face at the data-hull edge, Sec.~\ref{sec:hull}) visibly mixes
restyled ultra-black glyphs with lighter template fallbacks --- the
defect-tolerant pipeline providing critic-mediated functional recovery at
the cost of visible style consistency: evidence of fallback
functionality, not of a coherent type family on this donor.}
\label{fig:showcase}
\end{figure}

\paragraph{End-to-end generation.} A nine-prompt showcase run (one
open-vocabulary prompt per script, retrieval-selected donors) produced
nine installable fonts (Fig.~\ref{fig:showcase}) with shaped-cluster
accuracy 0.998--1.000 on every script
and legibility fraction 0.82--1.00, restyling 101--681 template glyphs
per font (the remainder shaped-cluster-safe template outlines) in
12--136 minutes on a single L4.
Across the 57 shipped presets the serving critic reports cluster accuracy
0.981--1.000; per-glyph repair reverts a median of $\sim$8\% of glyphs to
the template on the hardest display donors, and those reverts are
concentrated in rare conjuncts where a fallback is visually preferable to
a defect.

\paragraph{Structural validation.} All 66 shipped TTFs (57 presets, nine
showcase fonts) pass the OpenType Sanitizer (\texttt{ots}) --- the
load-time gate Chromium and Firefox apply to web fonts. Layout inheritance
was verified cross-engine: on conjunct- and matra-heavy probe strings per
script, HarfBuzz and Apple CoreText each produce glyph-ID sequences
identical to the font's template for all 66 fonts, with no system-font
fallback. A full-closure audit (every glyph of every font against its
template: 80{,}915 glyphs, 54{,}812 anchors) quantifies the metric
policy's actual footprint, split by provenance: the 50{,}066
template-fallback glyphs carry exactly zero metric deltas, while over
the 30{,}849 accepted generated glyphs advance deltas are p50
$0.045$\,em, p95 $0.358$\,em, maximum $1.93$\,em (per-script p95
$0.17$--$1.11$\,em, heaviest under Gujarati and Devanagari display
donors), with \emph{zero} narrowed advances closure-wide;
accepted-glyph right-sidebearing $|\Delta|$ reaches p95 $0.142$\,em
(max $1.98$), outline-extrema p95 $0.193$ (max $0.889$); anchor
translations are
purely horizontal ($\Delta y$ identically zero on all 54{,}812 compared
anchors, $|\Delta x|$ p95 $0.089$\,em, max $0.44$) and track their
outline's shift. Defining the tail as a ${>}0.35$\,em advance or
right-sidebearing change, 1{,}617 accepted glyphs ($5.2\%$; dominated by
i-matra width variants under display donors) exceed it across 61 fonts;
the supplement ships the per-glyph tail list and a SHA-256 manifest of
artifacts, templates, and validation logs. These are policy-defined
widenings, not typographic safety bounds. A rendered-context audit of the
tail addresses the gap directly: of the 1{,}617 tail glyphs, 248 are
reachable in the shaped unit inventory (the rest are feature-context
width variants); rendering each reachable glyph in flanked shaped context
at 12/24/48\,ppem against its template as control finds \emph{zero}
clipping, \emph{zero} line-fit blowups (advance ratio ${>}2.5\times$),
and 191 raw adjacent-ink overlaps of which 187 also occur in the
template (legitimate tight fit); the remaining \emph{four}
generation-introduced overlap cases (in three fonts) were reverted to
their template outlines, the three fonts rehashed and revalidated
(\texttt{ots} 66/66 still passes), and the rendered audit re-run clean.
Rendered
auditing of the 1{,}369 feature-context-only variants ($84.7\%$ of the
tail, reachable only through GSUB lookup contexts outside the unit
inventory) remains open beyond the shaped-cluster critic and should
precede any artifact freeze for a future human evaluation.
FontBakery's universal
profile reports the same
metadata-policy findings on our fonts as on the professional templates
themselves, plus exactly one generation-introduced defect class: our OFL
rename step dropped font-specific name records (IDs ${\geq}256$) that the
\texttt{STAT} table still references in fonts built on
variable-font-instanced templates (Bengali, Kannada, Odia), yielding
dangling-reference warnings. The rename step now preserves those records,
and the shipped files were repaired (the dropped records
restored from their templates; only name records changed, so the
outline/layout audit above remains applicable). The metadata-repaired
release was rerun through \texttt{ots} and FontBakery: \texttt{ots}
passes 66/66, the \texttt{ttx}-roundtrip
failure class drops from 22 fonts to zero with no dangling
\texttt{STAT} references, and every remaining FontBakery finding class
also occurs on the professional template controls. This FontBakery
comparison was run on the metadata-repaired release and was \emph{not}
repeated after the four outline reverts above: those three fonts were
re-validated with \texttt{ots} and the rendered-context audit only, so
the FontBakery findings describe their pre-revert outlines (an outline
revert can affect only FontBakery's per-glyph outline checks for the
four glyphs concerned). A SHA-256 manifest of
the repaired artifacts, their templates, and the validation logs
accompanies the supplement.
Scope and versions: \texttt{ots} 9.2.0; HarfBuzz via uharfbuzz 0.56.0 and
Apple CoreText on macOS~26.5, three conjunct- and matra-heavy probe
strings per script; FontBakery 1.1.0. These checks cover sanitizer
acceptance, glyph-ID parity, and the metric/anchor deltas above; they do
\emph{not} audit
anchor-to-ink attachment quality, cross-glyph collision, or rasterized
appearance ---
the shaped-cluster critic (Sec.~\ref{sec:system}) audits rendered
clusters, but only through HarfBuzz.

\paragraph{Style axes.} Parametric post-hoc axes (outline offsetting for
weight, signed-area counter scaling, em-fill scaling) apply bounded
geometric restyling to finished fonts within calibrated safe ranges
(weight offset $-40$ to $+80$ font units, counter scale 0.8--1.2, em-fill
0.95--1.06) with zero shaped-cluster breaks on calibration probes,
providing controllable weight and proportion variation that the diffusion
model alone does not expose.

\section{Negative Results and Forensics}
\label{sec:negative}
A third of this project's GPU budget went to levers that failed. We report
them because the pattern they form --- and the forensic method that
untangled the worst of them --- is more transferable than any single
positive result.

\subsection{Six tested-and-unsuccessful lever families}
Five attacked the same wall --- content fidelity under aggressive
styling, especially ultra-heavy display donors; the sixth attacked the
memory cost of per-script routing.

\begin{figure}[t]
\centering
\includegraphics[width=0.9\linewidth]{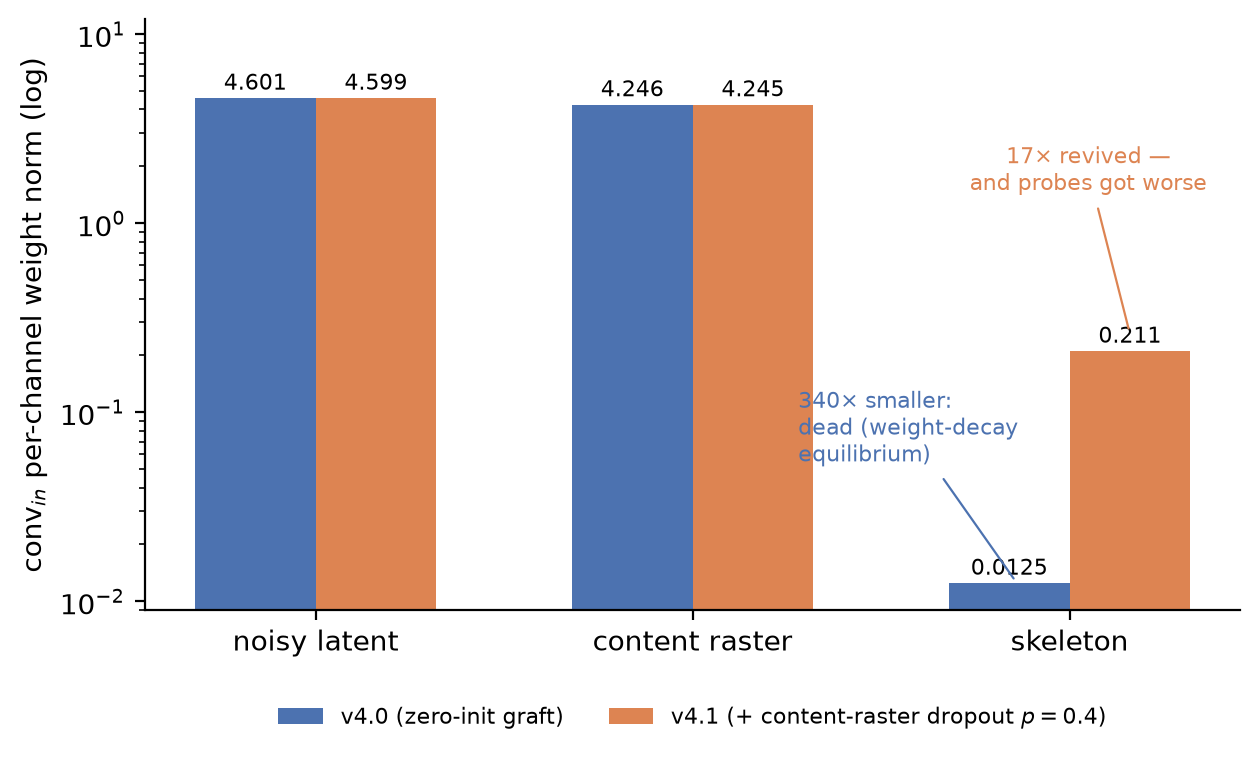}
\caption{Redundant conditioning dies. Per-channel \texttt{conv\_in}
weight norms after 25k graft steps. The zero-initialized skeleton channel
(v4.0) settles at weight-decay equilibrium because its signal is
computable from the adjacent raster channel. Content-raster dropout
(v4.1) makes the skeleton non-redundant and revives the pathway
$17\times$ --- and heavy-donor probes still got \emph{worse}, falsifying
the structure-deficit hypothesis rather than the mechanism.}
\label{fig:deadch}
\end{figure}

\textbf{(1) Auxiliary content CE on the denoised latent} (train-time
classifier pressure toward glyph identity): destroys global training ---
see the contamination study below.
\textbf{(2) Skeleton conditioning at \texttt{conv\_in}} (a Zhang-Suen
skeleton as a second content channel): the channel \emph{dies}. Because
the skeleton is a deterministic function of the raster channel beside it,
it carries zero marginal information; a zero-initialized channel whose
signal is redundant receives no sustained gradient while weight decay
pulls it to equilibrium --- final per-channel input norms 4.25 (raster)
vs 0.0125 (skeleton), a $340\times$ gap (Fig.~\ref{fig:deadch}).
ControlNet-style grafts work
precisely when the control input is \emph{not} computable from existing
inputs; a redundancy-breaking raster-dropout variant revived the pathway
($17\times$ norm recovery) but still failed gates.

\textbf{(3) Multi-scale content adapter} (FontDiffuser-style injection at
every down-block): the mechanism verifiably works --- adapter head norms
alive at all four scales, structure flows with the raster channel blanked
--- yet gates do not improve. Conditioning richness was not the binding
constraint.
\textbf{(4) Weight-domain target augmentation} (consistent
dilation/erosion of target+refs to extend the weight range): gates flat.
\textbf{(5) Self-supervised contrastive style loss} (predict the ref
embedding from the noised target; properly hinged per-sample): taxes
fidelity below bars, zero blackness gain.
\textbf{(6) Per-pack weighted sampling} (unify routing by re-weighting
old scripts to their pre-graft share): the first two runs collapsed ---
a verdict later \emph{voided} when both proved to carry the contaminating
aux loss of (1). A clean isolation (identical recipe minus the aux term)
trains healthily, and lands 2/7 gates: mean SSIM shift vs uniform
concatenation over the same window is $\approx 0$, but the dilution
\emph{relocates} to whichever scripts lose share (Malayalam, squeezed
from its 19\% natural share to 13\%, pays $-0.010$). Weighted sampling
is not toxic --- it is zero-sum: at this model capacity and window
length no share vector clears seven bars simultaneously, so routing
(Sec.~\ref{sec:training}) remains the architecture by controlled
evidence rather than by default.

Table~\ref{tab:levers} summarizes the campaign. The pattern that
argues against the conditioning hypothesis is the conjunction of the two
right-hand columns: in every row the mechanism check passes --- the
auxiliary head classifies, the revived channel carries signal, the
adapter heads contribute, the augmented densities reach 0.82, the
oversampled tail triples, the contrastive accuracy climbs --- yet the
extreme-donor fallback rate never improves on the 54\% baseline, and
five of eight variants pay for the attempt with a broken deva gate.

\begin{table}[t]
\centering
\small
\begin{tabular}{@{}llccc@{}}
\toprule
Lever & Attack surface & Mechanism & Deva & AlfaSlab fb \\
\midrule
v3.6 baseline & --- & --- & \textbf{0.7445} & 54\% \\
\midrule
Aux content CE & objective & \cmark & 0.553$^\dagger$ & ---$^\dagger$ \\
Skeleton ch. (v4.0) & input cond. & \xmark\ (dead) & 0.7366 & 56\% \\
+ raster drop (v4.1) & input cond. & \cmark & 0.7403 & 67\% \\
Multi-scale (v5.0) & decoder cond. & \cmark & 0.7363 & 59\% \\
Weight aug (v5.1) & data (synth) & \cmark & 0.7214 & 64\% \\
Oversample (v5.2) & data (balance) & \cmark & 0.7354 & 54\% \\
SCR unhinged (v5.3) & objective & \cmark & 0.6713 & 86\% \\
SCR hinged (v5.4) & objective & \cmark & 0.6882 & 53\% \\
\bottomrule
\end{tabular}
\caption{Six tested-and-unsuccessful lever families. \emph{Mechanism} = did the lever
verifiably do its job (channel alive, aug applied, loss learning)?
\emph{Deva} = SSIM on the deva test split under the protocol of record
(bar 0.7445). \emph{AlfaSlab fb} = template-fallback rate on an
out-of-hull ultra-black donor (baseline 54\%). Every working mechanism
either left the fallback rate unchanged or made it worse, while five of
eight broke the in-hull gate. $^\dagger$The aux-CE row predates the
protocol of record and the AlfaSlab probe (its contemporary deva baseline
was 0.729, so 0.553 is a 24\% collapse); it is also the one lever that
\emph{did} cut content errors (Lobster fallbacks $157\!\to\!48$) --- by
pulling outputs toward classifier centroids, i.e.\ it bought content by
destroying style. v4.0's mechanism
check \emph{failed} (skeleton channel norm $0.0125$, $340\times$ below
siblings) --- which is itself the finding.}
\label{tab:levers}
\end{table}

The convergent lesson: (1)--(5) attacked conditioning and objectives, but
the model was never \emph{asked} to draw glyphs beyond its target
distribution --- the failure is in the data hull (Sec.~\ref{sec:hull}),
and none of the conditioning levers we tested extrapolated a weight
regime absent from targets.

\subsection{The aux-loss contamination study}
\label{sec:contamination}
\begin{figure}[t]
\centering
\includegraphics[width=\linewidth]{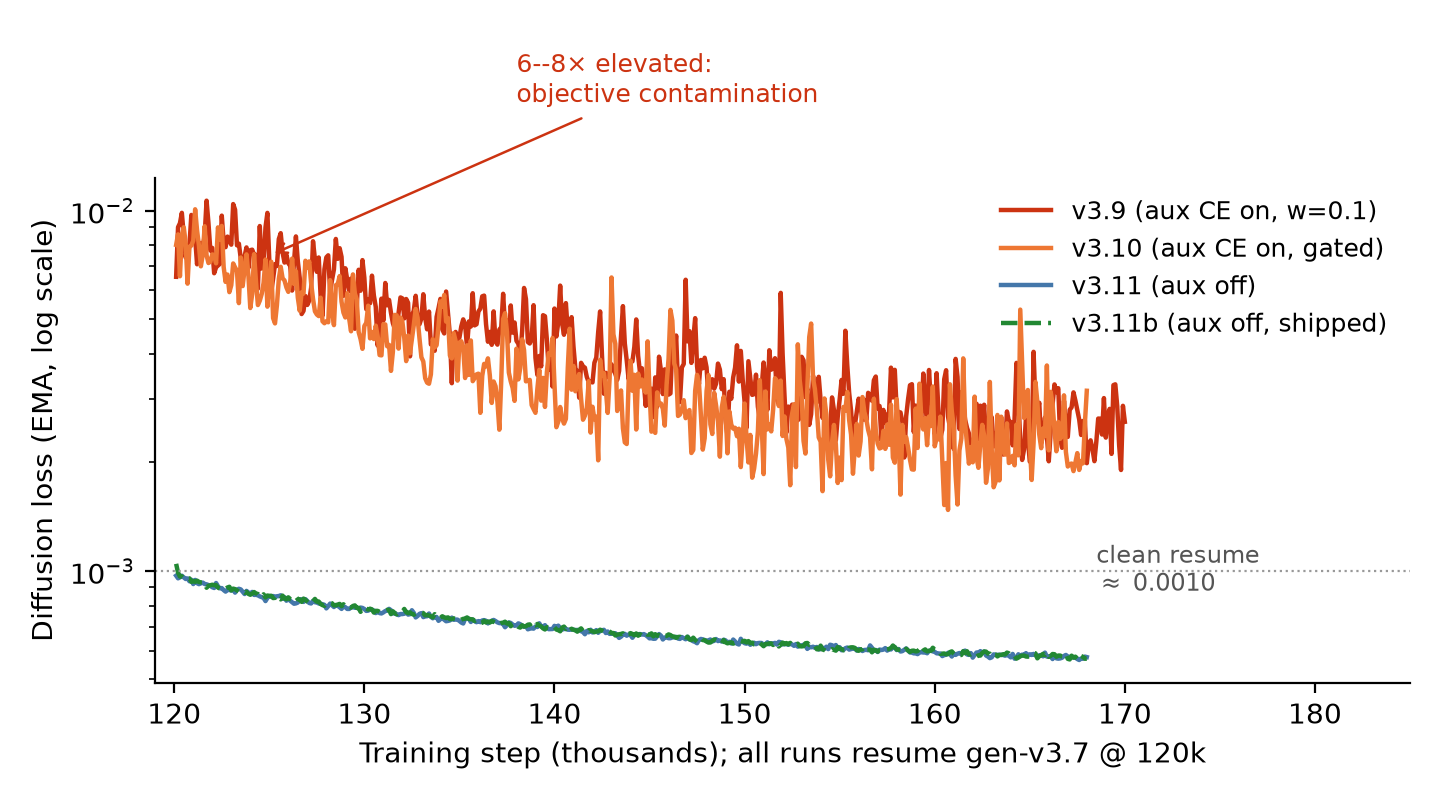}
\caption{Objective contamination has an instant loss fingerprint. Four
runs resume the same gen-v3.7 checkpoint at step 120k with the same LR
schedule; the only difference is a train-time content-identity CE term
(weight 0.1). Contaminated runs (red/orange) sit 6--8$\times$ above the
clean diffusion-loss regime from the first logged step and never rejoin
it; clean runs (blue/green) overlap almost exactly. Gates would not
reveal the failure until $\sim$7 GPU-hours later.}
\label{fig:loss}
\end{figure}

A train-time content-identity loss (weight 0.1) --- previously only a
serve-time filter in passing runs --- was carried into four consecutive
training launches through a misreconstructed launch configuration. All
four collapsed with one signature: \emph{uniform} cross-script SSIM
collapse (0.13--0.19 below bars on all seven scripts at once),
systematic boldening (generated/target ink ratio $0.963 \to 1.036$), and
style washing, while validation panels, reconstruction grids, and the
loss curve all looked healthy. Controlled reruns cleared three plausible
culprits: a sampler-free run collapsed identically (not the new
sampler); a byte-exact rerun of a passing recipe collapsed (not the
window or schedule); re-evaluating unchanged checkpoints reproduced
their historical scores to four decimals (not environment drift). The
experiment tracker settled it: diffing the logged \texttt{config} of
passing and failing runs isolated a single delta
(\texttt{content\_ckpt}), and the diffusion-term loss at the resume step
separates the regimes instantly (Fig.~\ref{fig:loss}: 0.0010 clean vs
0.0066--0.0090 polluted --- the gap \emph{is} the aux term). A clean
relaunch passed its gates on the first try.

We draw three operational lessons. (i)~Never add a train-time loss term
absent from the last passing run's logged config; diff configs
\emph{before} launching. (ii)~Uniform cross-script collapse with healthy
panels and systematic boldening indicates objective-function
contamination, not data, sampling, or environment. (iii)~The experiment
tracker's per-run config and environment snapshot are the forensic record
of truth; a launch reconstructed from memory is not a record.

\subsection{The data-hull boundary}
\label{sec:hull}
\begin{figure}[t]
\centering
\includegraphics[width=\linewidth]{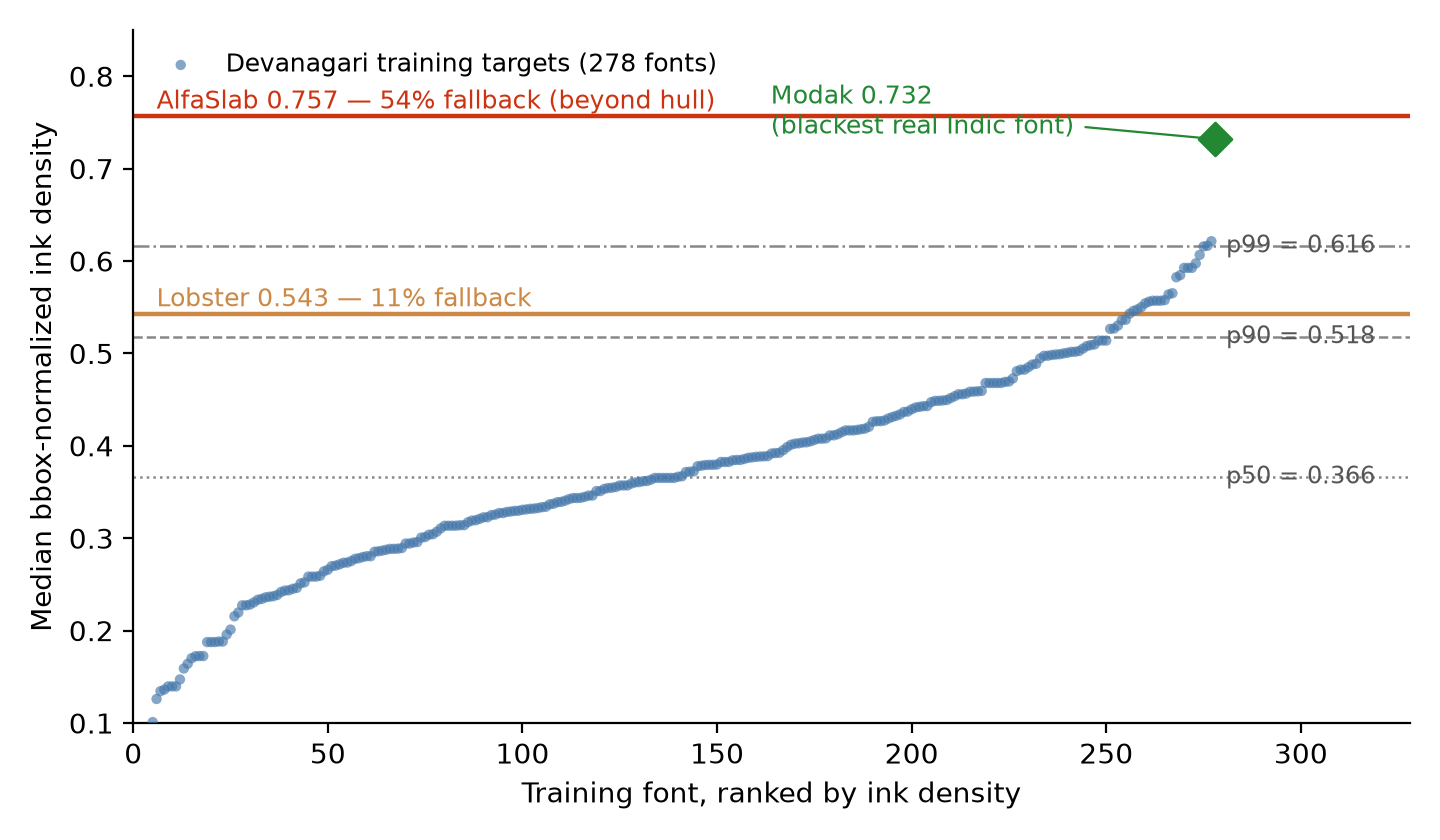}
\caption{The data hull. Per-font median bbox-normalized ink density of
all 278 Devanagari training faces, ranked. The beyond-hull probe donor
AlfaSlab (0.757) is blacker than every training target including Modak
(0.732); its 54\% content-gate fallback rate was immovable under all six
unsuccessful levers. Lobster (0.543, $\approx$p90) sits inside the hull and
restyles with 11\% fallback.}
\label{fig:hull}
\end{figure}

Measuring per-font median bbox-normalized ink density over all 278
Devanagari training faces (complete families; multi-weight families
contribute several faces; Fig.~\ref{fig:hull}): p50 0.366, p90 0.518,
p99 0.616, maximum 0.732 (Modak, the blackest real Indic font in the
corpus). Under the same metric the ultra-black Latin donor AlfaSlab
measures 0.757 --- beyond \emph{every} training target --- and exhibits a
structural 54--59\% content-gate fallback rate under every lever in this
section; Lobster at 0.543 ($\approx$p90) mostly works (11\% fallback). The
serving system therefore ships the boundary rather than pretending to
cross it: a donor ink-density precheck bands each request (normal /
heavy / hull-edge / beyond-hull) with calibrated expected-fallback
warnings. Under this reading --- and limited to the architecture, critic,
and levers we tested --- crossing the hull points to targets that do
not exist in open-license Indic typography, i.e.\ commissioned glyph
sheets, rather than to the modeling levers we tried. We frame the hull as the
best-supported \emph{hypothesis}, not a demonstrated cause: the evidence
is a single scalar (ink density) and two Latin probe donors; AlfaSlab
(0.757) exceeds the blackest training face (0.732) only modestly; and
explanations at other levels --- backbone capacity, critic calibration,
tile resolution, or the Latin-to-Indic donor gap itself --- are not
excluded. A density-banded multi-donor study would be needed to map the
boundary's actual shape.

\section{Limitations}
\label{sec:limitations}
\srijika{} inherits the \emph{data hull}: donors whose ink density exceeds
the blackest training faces (Sec.~\ref{sec:hull}) degrade to
template-heavy output; under our best-supported hypothesis the remedy is
data acquisition rather than the modeling levers we tested, though
modeling-side explanations (capacity, critic calibration, tile
resolution, the donor-script gap) are not excluded (Sec.~\ref{sec:hull}).
Style transfer is glyph-local: donor identities carried by layout-scale
features --- extreme width (expanded faces), connected-script joins,
baseline dance --- transfer weakly even when per-glyph quality is high
(donor match cosine 0.24--0.41 on such donors vs 0.61--0.99 on
weight/contrast-driven styles), because six reference tiles cannot encode
what is not visible at tile scale; a \emph{library} of structurally
diverse audited templates per script (wide, condensed, high-contrast),
matched to the donor before restyling, could recover some of this ---
at the cost of re-grafting, since the content channel's distribution is
tied to the training template. Per-script routing trades memory for
regression-safety; a unified single model remains open (weighted
sampling trains cleanly but reallocates dilution zero-sum,
Sec.~\ref{sec:negative}; larger capacity, continual-learning
regularization such as EWC, per-script adapters over a frozen shared
backbone, or mixture-of-experts capacity are the natural next levers,
none yet run through our falsification protocol). Generated fonts reuse the template's
kerning and mark-attachment data (subject to the anchor-translation
policy of Sec.~\ref{sec:system}) but ship unhinted, and parametric axes
operate on outlines rather than variable-font deltas. Our evaluation has
three further gaps, each named plainly. First, our gates
measure fidelity to \emph{held-out real fonts} under oracle references;
they isolate the reconstruction component and neither bound nor
measure open-vocabulary prompt quality, for which we rely on designer
review rounds. The prompt interface (retrieval) is validated only by
an attribute-label Spearman of 0.50 (Sec.~\ref{sec:system}); a
retrieval evaluation against designer-labeled prompts and a paired human
preference study (generated vs.\ geometric vs.\ template fonts, for
which the non-learning arms are already built, Appendix~\ref{app:perface})
are the missing evidence and remain future work. Second, all baselines are no-learning
references; a learned few-shot generator (FontDiffuser, MX-Font, CF-Font,
DG-Font) trained on the same pair packs and scored under the same gate
protocol is the comparison that would make our claims about which levers
bind falsifiable. Third, on heavy display donors the shipped output is a
functional font, not a coherent type family: restyled and fallback
glyphs visibly mix (Fig.~\ref{fig:showcase}, Tamil row), and while we
report fallback rates we do not yet report a within-font coherence
statistic (e.g., stroke-weight dispersion) per preset.

\section{Conclusion}
\srijika{} demonstrates prompt-guided restyling of
template-layout-reusing fonts across nine Brahmic scripts, shipped as
installable TTFs --- not by solving font
generation in general, but by choosing a formulation in which
layout-table structure is reused from existing professional
engineering --- subject to documented metric and anchor-coordinate
transformations --- and the learned component is confined to what diffusion
models are demonstrably good at: local visual restyling under reference
conditioning. The system-level levers that mattered (genre-matched
splits, graft-with-routing, defect-tolerant serving) are all data and
protocol; the model-level levers we tried mostly did not, and we
documented why. We hope the negative-results catalogue and the data-hull
analysis save others the GPU cycles, and that the shipped nine-script
system --- with its 57 presets and open evaluation protocol --- becomes a
baseline for typographic generation in underserved scripts.

\section*{Ethics and Licensing}
Every training font, template, retrieval-index entry, and donor in this
work is distributed under the SIL Open Font License or Apache~2.0.
Generated fonts are derivatives of their template font: we retain the
template's license, rename per OFL Reserved Font Name requirements, and
embed attribution to the template family in the output's name table.
No proprietary or scraped fonts were used at any stage. Generated fonts
are synthetic derivatives and should not be represented as the work of
the original designers; conversely, the system is designed to expand ---
not displace --- thin typographic ecosystems, and we release only
open-licensed outputs.

\section*{Code and Data Availability}
\emph{Included with this version} (arXiv ancillary files): the frozen
\texttt{benchmark\_v1} manifest; the per-face, per-seed evaluation CSV
for every method (with the six reference-tile indices per face); the
per-glyph metric-tail list and its rendered-context audit; the SHA-256
manifest of all 66 shipped fonts, the nine geometric-baseline fonts,
their templates, and the validation logs; and the human-evaluation
protocol sketch. \emph{Public now}: the \lipika{} recognizer and
retrieval code (\url{https://github.com/Loopdesk-AI/lipika}), recognizer
weights (\url{https://huggingface.co/loopdesk-ai/lipika}), and the
recognizer evaluation set
(\url{https://huggingface.co/datasets/loopdesk-ai/lipika-eval}).
\emph{Not included with this version}: the \srijika{} generator,
serving pipeline, and evaluation implementation (including the
geometric baseline), which are scheduled for release in a subsequent
revision; and the generated font files themselves, for which the
ancillary manifest carries hashes only.

\appendix
\section{Gate protocol details}
\label{app:gate}
Each script has a frozen pair pack (tag \texttt{\{script\}\_v1}) whose
test split holds out complete families. Evaluation renders the script's
unit inventory for each held-out family, samples the model at guidance
3.0 with six reference tiles drawn from that family, and scores SSIM
between generated and ground-truth $128{\times}128$ tiles; the report
mean and 25th percentile pool all generated tiles of the held-out faces
(Table~\ref{tab:gates}; per-face breakdown in
Appendix~\ref{app:perface}). Templates are pinned per script (e.g.\ Baloo
Tamma~2 for Kannada, Mukta Mahee for Gurmukhi, Noto Sans Oriya wght400
for Odia). A run may ship for a script only if it clears both the gate
and the script's inherited bar; evaluation artifacts (per-family scores,
recon grids) are archived with the checkpoint.

\section{Per-face gate scores and template baseline}
\label{app:perface}
Each held-out face contributes exactly 64 scored tiles, so the pooled
means of Table~\ref{tab:gates} coincide with per-face macro-averages;
per-face p25, however, is not recoverable from pooled numbers, and tiles
within a face are correlated, so the face-level rows in
Tables~\ref{tab:perface}--\ref{tab:perfaceb} are the honest unit of
display --- and since the weights of one design are themselves
correlated, the independent unit is the \emph{family} (counts and
family-level macro-averages in Table~\ref{tab:gates}). \emph{Tpl.\ SSIM}
is the template-copy baseline on the same targets and \emph{Morph} the
density-matched morphology reference --- a target-oracle diagnostic that
reads each scored target's ink coverage, not a reference-conditioned
baseline (Sec.~\ref{sec:results}); per-face values for the reference-only
geometric baseline and its gap closure are in the supplemental CSV, with
family-level values in Table~\ref{tab:family}.
\emph{Style ratio} is computed in \lipika{} embedding space: for each
held-out face, the six reference tiles that condition the generator are
embedded by the \lipika{} recognizer backbone (the generator's own frozen
style encoder, Sec.~\ref{sec:system}) and L2-normalized, and their
normalized mean $c$ is the face's style centroid; with
$\bar{s}_{\mathrm{gen}}$ the mean cosine similarity of generated-tile
embeddings to $c$, and $\bar{s}_{\mathrm{real}}$ the same for the face's
real tiles (the scored targets), the reported ratio is
$\bar{s}_{\mathrm{gen}}/\max(\bar{s}_{\mathrm{real}},10^{-6})$. It is a
\emph{normalized absolute similarity} to the reference centroid --- not a
displacement measure: it does not subtract the template's score, so a
value of 0.5 does not mean half of a style transformation
occurred.\footnote{The implementation of this column --- the cosine ratio
above, computed by the same evaluation code --- is identical across all
evaluation rounds of this project, and face-level values reproduce
bit-identically between rounds; an earlier draft's textual description of
the column (as an ink-density-recovery measure) was a documentation
error, corrected as of this revision.} It is
unclipped and can exceed~1; the denominator is far from degenerate
($\bar{s}_{\mathrm{real}}\geq 0.78$ across the 56 faces). Unrestyled
template tiles score 0.06--0.17 pooled per script on the same statistic
(per-face range $-0.01$ to $0.45$) --- template glyphs
already share much of any face's embedding signature, so generated
ratios should be read against that nonzero floor, not against zero.
A paired, template-relative variant --- the \emph{\lipika{}
similarity-gap closure}
$p=(\bar{s}_{\mathrm{gen}}-\bar{s}_{\mathrm{tpl}})/(\bar{s}_{\mathrm{real}}-\bar{s}_{\mathrm{tpl}})$,
with $\bar{s}_{\mathrm{tpl}}$ the template tiles' mean cosine to the same
centroid, is reported per family in Table~\ref{tab:family} (unclipped;
face-level denominators 0.52--0.95). Both remain \emph{internal}
scores: the embedding is the generator's own frozen style encoder, a
family-identity recognizer whose training corpus includes the held-out
families (Sec.~\ref{sec:results}), so these are same-model measurements
by a judge trained to recognize the targets --- not independent style
validation.
\paragraph{Geometric-baseline specification.} The reference-only
geometric baseline of Sec.~\ref{sec:results} is estimated as follows.
\emph{Pairing:} for each held-out face, the six donor reference tiles
and the template tiles of the \emph{same six unit indices} (the
rng-fixed indices exported in the manifest) are paired. \emph{Tile
conventions:} all tiles are the pack's 128\,px renders (antialiased
greyscale, ink white-on-black, mapped to $[0,1]$ with ink $=1$);
reference tiles, stored on the recognizer's $224$\,px canvas, are
cropped back to the centered $128$\,px frame before any statistic.
\emph{Statistics:} per tile, the ink bounding box is computed on a
$0.5$ binarization (first/last ink row and column) and ink coverage is
the mean of the $[0,1]$ tile. \emph{Estimation:} $s_x$ is the ratio of
the mean reference bbox width to the mean template bbox width (means
over the six pairs), $s_y$ likewise for heights; each is clamped to
$[0.8, 1.25]$ after the ratio (no re-fitting). The scale is applied
about the tile center by bilinear resampling with zero (background)
padding. Weight is one fractional morphological radius on a nine-step
ladder $r \in \{-4,\dots,4\}$ (min/max-pool dilation/erosion of ink):
the ladder is applied to the \emph{scaled} template reference tiles,
mean ink per radius is monotone in $r$, and the references' mean ink
coverage is bracketed by two adjacent radii and linearly interpolated
(lower bracket on ties; interpolation weight clamped to $[0,1]$; ladder
ends clamp). \emph{Order and scope:} scale first, then weight; exactly
one $(s_x, s_y, r)$ tuple per held-out face, estimated from that face's
six references only --- no scored-target information --- and applied
unchanged to all of the face's scored template tiles. \emph{Scoring:}
the transformed tiles are scored in tile space by the same SSIM and
\lipika{} embedding path as every other method; they are not
vectorized or assembled into a font for the gate comparison --- we
therefore call it a coherent \emph{tile-space} geometric transform. We
additionally assembled one geometric font per script through the same
vectorization, metric, and assembly pipeline as \srijika{} (donor = the
gate protocol's first held-out face, identical six references and
parameters): all nine pass \texttt{ots} with zero blank or failed
glyphs, their SHA-256 hashes join the supplement manifest, and they
stand ready as a non-learning arm for a future human evaluation (a
protocol sketch ships in the supplement). The
evaluation implementation is not included with this version and is
scheduled for a later release (Code and Data Availability).
Re-running the frozen evaluation end-to-end reproduced pooled means
bit-exactly on four of nine scripts and within $7{\times}10^{-5}$ on the
rest (GPU-level nondeterminism across workers); these tables report the
re-run. Every evaluation report also exports, per face, the exact six
reference-tile indices used for conditioning and the raw
$\bar{s}_{\mathrm{gen}}/\bar{s}_{\mathrm{tpl}}/\bar{s}_{\mathrm{morph}}/\bar{s}_{\mathrm{geo}}/\bar{s}_{\mathrm{real}}$
cosines, and the supplemental CSV carries all per-face, per-seed values,
including the geometric baseline and the Odia style-OOD condition of
Table~\ref{tab:alkatra}.

\begin{sidewaystable}[p]
\centering
\footnotesize
\caption{Controlled two-strata benchmark (\texttt{benchmark\_v1}): one
model (gen-bench-v1, 168k steps from scratch, all nine scripts) with
every ID/OOD/validation family excluded from diffusion pair-pack
training --- these families were \emph{not} excluded from \lipika{}
encoder or retrieval-index training --- evaluated on both strata under
identical weights, references, and baselines. Columns as in
Table~\ref{tab:family}; \emph{Str.}\ marks genre-matched ID vs.\
style-OOD hold-outs (OOD family names bold) as fixed in the frozen
manifest, which can differ from the shipped models' earlier test splits
(Akaya Kanadaka, the display cut of the shipped Kannada test split in
Tables~\ref{tab:family}--\ref{tab:perfaceb}, is Kannada's style-OOD
family here); \emph{Closure seeds} gives the generated \lipika{}
similarity-gap closure at diffusion seeds 7/1/2 (inference sampling
only; one training run). Alkatra posts high closure in both its script
renderings (0.935 Bengali, 0.853 Odia --- correlated observations of one
held-out family) while the other seven unique OOD families close a
median 0.040. Nine OOD script-family rows, eight unique OOD families;
all statistics descriptive (1--5 faces per OOD family).}
\label{tab:bench}
\footnotesize
\setlength{\tabcolsep}{5.0pt}
\begin{tabular}{llcrcccccccl}
\toprule
& & & & \multicolumn{4}{c}{SSIM} & \multicolumn{3}{c}{\lipika{} closure} & \\
\cmidrule(lr){5-8}\cmidrule(lr){9-11}
Script & Held-out family & Str. & Faces & Gen & Tpl & Morph & Geom &
Gen & Geo & Morph & Closure seeds \\
\midrule
Devanagari & Glegoo & ID & 2 & 0.7001 & 0.7182 & 0.7133 & 0.7206 & 0.010 & 0.029 & 0.023 & 0.010/0.000/0.005 \\
 & Karma & ID & 5 & 0.7644 & 0.8178 & 0.8181 & 0.7881 & 0.141 & 0.030 & 0.014 & 0.141/0.149/0.140 \\
 & Khand & ID & 5 & 0.7058 & 0.7940 & 0.7972 & 0.7925 & 0.195 & 0.005 & 0.011 & 0.195/0.196/0.194 \\
 & \textbf{Modak} & OOD & 1 & 0.6924 & 0.7449 & 0.7855 & 0.7412 & 0.062 & 0.018 & 0.018 & 0.062/0.054/0.064 \\
Tamil & Karla Tamil Upright & ID & 2 & 0.7508 & 0.7601 & 0.7628 & 0.6945 & 0.412 & $-$0.036 & $-$0.017 & 0.412/0.349/0.378 \\
 & Meera Inimai & ID & 1 & 0.7731 & 0.7592 & 0.7635 & 0.6891 & 0.989 & $-$0.004 & 0.006 & 0.989/0.983/0.990 \\
 & \textbf{Coiny} & OOD & 1 & 0.6935 & 0.6894 & 0.6746 & 0.6684 & $-$0.019 & $-$0.005 & $-$0.016 & $-$0.019/$-$0.025/$-$0.021 \\
Bengali & Mina & ID & 2 & 0.7649 & 0.7918 & 0.7851 & 0.7464 & 0.045 & 0.029 & 0.020 & 0.045/0.039/0.049 \\
 & Tiro Bangla & ID & 2 & 0.7550 & 0.7681 & 0.7687 & 0.7754 & 0.050 & 0.009 & 0.001 & 0.050/0.055/0.052 \\
 & \textbf{Alkatra} & OOD & 5 & 0.7242 & 0.7470 & 0.7429 & 0.7505 & 0.935 & 0.006 & 0.011 & 0.935/0.936/0.938 \\
Telugu & Chathura & ID & 5 & 0.7546 & 0.7729 & 0.7980 & 0.8425 & 0.114 & $-$0.004 & $-$0.001 & 0.114/0.120/0.118 \\
 & Ravi Prakash & ID & 1 & 0.7174 & 0.7853 & 0.7820 & 0.8097 & 0.083 & $-$0.001 & $-$0.000 & 0.083/0.077/0.091 \\
 & Sirivennela & ID & 1 & 0.7440 & 0.7648 & 0.7829 & 0.8233 & 0.020 & 0.040 & 0.033 & 0.020/0.010/0.029 \\
 & Tenali Ramakrishna & ID & 1 & 0.7364 & 0.8068 & 0.8116 & 0.8135 & $-$0.151 & 0.012 & 0.019 & $-$0.151/$-$0.176/$-$0.122 \\
 & \textbf{Akaya Telivigala} & OOD & 1 & 0.7234 & 0.7531 & 0.7489 & 0.7547 & $-$0.034 & $-$0.000 & 0.005 & $-$0.034/$-$0.028/$-$0.030 \\
Kannada & Noto Serif Kannada & ID & 10 & 0.7528 & 0.7587 & 0.7589 & 0.7482 & 0.452 & 0.119 & 0.093 & 0.452/0.422/0.427 \\
 & \textbf{Akaya Kanadaka} & OOD & 1 & 0.7019 & 0.7130 & 0.7110 & 0.7467 & 0.019 & $-$0.008 & 0.002 & 0.019/0.026/0.019 \\
Malayalam & Lohit Malayalam & ID & 1 & 0.7205 & 0.7304 & 0.7245 & 0.7057 & 0.148 & $-$0.005 & $-$0.021 & 0.148/0.127/0.093 \\
 & RIT Panmana & ID & 1 & 0.8018 & 0.8150 & 0.8192 & 0.7993 & 0.201 & 0.218 & 0.014 & 0.201/0.218/0.190 \\
 & RIT TN Joy & ID & 3 & 0.7156 & 0.7943 & 0.7947 & 0.8196 & 0.272 & $-$0.011 & 0.023 & 0.272/0.273/0.294 \\
 & RIT Uroob & ID & 2 & 0.6936 & 0.7713 & 0.7730 & 0.8149 & 0.133 & $-$0.002 & $-$0.004 & 0.133/0.118/0.133 \\
 & \textbf{Chilanka} & OOD & 1 & 0.6966 & 0.7688 & 0.7688 & 0.7457 & 0.055 & $-$0.015 & $-$0.002 & 0.055/0.068/0.055 \\
Gujarati & Kumar One Outline & ID & 1 & 0.6813 & 0.7068 & 0.7038 & 0.7022 & 0.004 & $-$0.004 & 0.001 & 0.004/0.003/0.007 \\
 & Mogra & ID & 1 & 0.7192 & 0.7453 & 0.7361 & 0.7467 & $-$0.005 & 0.010 & 0.016 & $-$0.005/$-$0.005/$-$0.005 \\
 & \textbf{Shrikhand} & OOD & 1 & 0.6941 & 0.7118 & 0.7097 & 0.7093 & 0.039 & 0.005 & 0.015 & 0.039/0.036/0.041 \\
Gurmukhi & Braah One & ID & 1 & 0.7767 & 0.7618 & 0.7676 & 0.7380 & $-$0.036 & 0.024 & 0.054 & $-$0.036/$-$0.038/$-$0.037 \\
 & Tiro Gurmukhi & ID & 2 & 0.8326 & 0.8234 & 0.8261 & 0.7886 & 0.219 & 0.005 & 0.003 & 0.219/0.217/0.234 \\
 & \textbf{Langar} & OOD & 1 & 0.7470 & 0.7663 & 0.7532 & 0.7524 & 0.040 & 0.003 & $-$0.002 & 0.040/0.034/0.041 \\
Odia & Lohit Oriya & ID & 1 & 0.7218 & 0.7619 & 0.7628 & 0.7325 & 0.121 & 0.020 & $-$0.003 & 0.121/0.121/0.129 \\
 & Noto Serif Oriya & ID & 5 & 0.7193 & 0.7893 & 0.7873 & 0.7546 & 0.034 & 0.004 & 0.007 & 0.034/0.037/0.033 \\
 & \textbf{Alkatra} & OOD & 5 & 0.6418 & 0.6989 & 0.6911 & 0.7106 & 0.853 & 0.010 & 0.008 & 0.853/0.831/0.854 \\
\bottomrule
\end{tabular}
\end{sidewaystable}

\begin{table}[p]
\centering
\footnotesize
\caption{Family-level results, all methods (test split, guidance 3).
Faces are averaged within each held-out family first (the family is the
independent unit; $n{=}2$--$4$ families per script, so all statistics are
descriptive). SSIM columns: generated, template copy, the
target-oracle morphology diagnostic, and the coherent reference-only
geometric baseline (one global ink-bbox scale and weight radius
estimated from the six references). Style columns: mean \lipika{} cosine
to the face's six-reference centroid for generated, template, geometric,
and real target tiles ($\bar{s}_{\mathrm{morph}}$ tracks
$\bar{s}_{\mathrm{tpl}}$ within 0.08 everywhere and is in the CSV), and
the \lipika{} similarity-gap closures
$p=(\bar{s}_{\mathrm{gen}}-\bar{s}_{\mathrm{tpl}})/(\bar{s}_{\mathrm{real}}-\bar{s}_{\mathrm{tpl}})$
and $p_{\mathrm{geo}}$ (same formula for the geometric baseline; both
unclipped, negative when the method lands farther from the centroid
than the template). Generated style exceeds the template's on 18 of 23
families, but the median closure is 0.047 (IQR 0.016--0.133; 8 families
above 0.10, 3 above 0.25, 5 negative) --- the typical effect is small
and the large closures concentrate in a few families. The geometric
baseline is the structural mirror image: higher SSIM than generation on
17 of 23 families, but median closure 0.006 with none above 0.25.
Sampling-seed sensitivity: rerunning all nine gates with two additional
diffusion seeds moves pooled per-script SSIM by at most $0.006$ and
family macros by at most $0.009$ (median $0.004$); the closure medians
are 0.047/0.054/0.047 across seeds with unchanged threshold counts and a
single near-zero sign flip (Lohit Malayalam); template and
morphology columns are seed-invariant. Values are the
protocol-of-record seed; a machine-readable per-face, per-seed CSV
(including the six reference-tile indices per face) accompanies the
submission.}
\label{tab:family}
\setlength{\tabcolsep}{2.6pt}
\begin{tabular}{llrcccccccccc}
\toprule
& & & \multicolumn{4}{c}{SSIM} & \multicolumn{6}{c}{\lipika{} style} \\
\cmidrule(lr){4-7}\cmidrule(lr){8-13}
Script & Held-out family & Faces & Gen & Tpl & Morph & Geom &
$\bar{s}_{\mathrm{gen}}$ & $\bar{s}_{\mathrm{tpl}}$ &
$\bar{s}_{\mathrm{geo}}$ &
$\bar{s}_{\mathrm{real}}$ & $p$ & $p_{\mathrm{geo}}$ \\
\midrule
Devanagari & Glegoo & 2 & 0.6953 & 0.7182 & 0.7133 & 0.7206 & 0.081 & 0.105 & 0.127 & 0.833 & $-$0.033 & 0.029 \\
 & Karma & 5 & 0.7553 & 0.8178 & 0.8181 & 0.7881 & 0.143 & 0.008 & 0.036 & 0.917 & 0.148 & 0.030 \\
 & Khand & 5 & 0.7608 & 0.7940 & 0.7972 & 0.7925 & 0.134 & 0.070 & 0.074 & 0.898 & 0.077 & 0.005 \\
Tamil & Karla Tamil Upright & 2 & 0.7361 & 0.7615 & 0.7644 & 0.6910 & 0.477 & 0.168 & 0.143 & 0.941 & 0.400 & $-$0.032 \\
 & Meera Inimai & 1 & 0.7988 & 0.7563 & 0.7594 & 0.6932 & 0.894 & 0.133 & 0.125 & 0.910 & 0.979 & $-$0.010 \\
Bengali & Mina & 2 & 0.7677 & 0.7946 & 0.7879 & 0.7576 & 0.158 & 0.104 & 0.127 & 0.918 & 0.066 & 0.028 \\
 & Tiro Bangla & 2 & 0.7698 & 0.7674 & 0.7682 & 0.7730 & 0.100 & 0.049 & 0.062 & 0.914 & 0.059 & 0.015 \\
Telugu & Chathura & 5 & 0.7840 & 0.7739 & 0.7985 & 0.8435 & 0.204 & 0.100 & 0.095 & 0.979 & 0.118 & $-$0.006 \\
 & Ravi Prakash & 1 & 0.7377 & 0.7838 & 0.7802 & 0.8159 & 0.118 & 0.132 & 0.131 & 0.962 & $-$0.017 & $-$0.001 \\
 & Sirivennela & 1 & 0.7515 & 0.7778 & 0.7938 & 0.8304 & 0.143 & 0.120 & 0.142 & 0.985 & 0.027 & 0.026 \\
 & Tenali Ramakrishna & 1 & 0.7436 & 0.8069 & 0.8113 & 0.8211 & 0.362 & 0.414 & 0.430 & 0.929 & $-$0.101 & 0.032 \\
Kannada & Akaya Kanadaka & 1 & 0.7127 & 0.7130 & 0.7110 & 0.7467 & 0.125 & 0.102 & 0.096 & 0.940 & 0.027 & $-$0.008 \\
 & Noto Serif Kannada & 10 & 0.6833 & 0.7587 & 0.7589 & 0.7482 & 0.264 & 0.117 & 0.220 & 0.981 & 0.170 & 0.119 \\
Malayalam & Lohit Malayalam & 1 & 0.6996 & 0.7304 & 0.7245 & 0.7057 & 0.208 & 0.191 & 0.188 & 0.946 & 0.023 & $-$0.005 \\
 & RIT Panmana & 1 & 0.7657 & 0.8150 & 0.8192 & 0.7993 & 0.346 & 0.249 & 0.374 & 0.821 & 0.170 & 0.218 \\
 & RIT TN Joy & 3 & 0.7285 & 0.7943 & 0.7947 & 0.8196 & 0.312 & 0.123 & 0.115 & 0.839 & 0.264 & $-$0.011 \\
 & RIT Uroob & 2 & 0.7649 & 0.7713 & 0.7730 & 0.8149 & 0.095 & 0.002 & 0.000 & 0.886 & 0.105 & $-$0.002 \\
Gujarati & Kumar One Outline & 1 & 0.6820 & 0.7068 & 0.7038 & 0.7022 & 0.138 & 0.130 & 0.127 & 0.996 & 0.009 & $-$0.004 \\
 & Mogra & 1 & 0.7218 & 0.7453 & 0.7361 & 0.7467 & 0.151 & 0.156 & 0.164 & 0.981 & $-$0.006 & 0.010 \\
Gurmukhi & Braah One & 1 & 0.7536 & 0.7618 & 0.7676 & 0.7380 & 0.184 & 0.193 & 0.211 & 0.946 & $-$0.012 & 0.024 \\
 & Tiro Gurmukhi & 2 & 0.8023 & 0.8253 & 0.8278 & 0.7914 & 0.134 & 0.095 & 0.100 & 0.936 & 0.047 & 0.006 \\
Odia & Lohit Oriya & 1 & 0.7290 & 0.7619 & 0.7628 & 0.7325 & 0.205 & 0.188 & 0.201 & 0.847 & 0.026 & 0.020 \\
 & Noto Serif Oriya & 5 & 0.7689 & 0.7917 & 0.7903 & 0.7608 & 0.164 & 0.143 & 0.148 & 0.971 & 0.025 & 0.006 \\
\bottomrule
\end{tabular}
\end{table}

\begin{table}[p]
\centering
\small
\caption{Per-face gate scores (test split, guidance 3): the five
originally shipped scripts (gen-v3.7).}
\label{tab:perface}
\begin{tabular}{llccccc}
\toprule
Script & Held-out face & SSIM & SSIM p25 & Tpl.\ SSIM & Morph & Style ratio \\
\midrule
Devanagari & Glegoo Bold & 0.6699 & 0.6361 & 0.7057 & 0.6966 & 0.075 \\
 & Glegoo & 0.7206 & 0.6652 & 0.7308 & 0.7300 & 0.123 \\
 & Karma Bold & 0.7273 & 0.6699 & 0.8109 & 0.8012 & 0.088 \\
 & Karma Light & 0.7793 & 0.7367 & 0.8268 & 0.8337 & 0.273 \\
 & Karma Medium & 0.7805 & 0.7434 & 0.8287 & 0.8304 & 0.163 \\
 & Karma & 0.7549 & 0.7087 & 0.8124 & 0.8167 & 0.174 \\
 & Karma SemiBold & 0.7346 & 0.6843 & 0.8102 & 0.8082 & 0.083 \\
 & Khand Bold & 0.7204 & 0.6676 & 0.7756 & 0.7855 & 0.101 \\
 & Khand Light & 0.7798 & 0.7448 & 0.7894 & 0.7957 & 0.307 \\
 & Khand Medium & 0.7686 & 0.7272 & 0.8019 & 0.8037 & 0.135 \\
 & Khand & 0.7865 & 0.7401 & 0.8035 & 0.8078 & 0.108 \\
 & Khand SemiBold & 0.7488 & 0.7050 & 0.7993 & 0.7934 & 0.104 \\
Tamil & Karla Tamil Upright Bold & 0.7256 & 0.6707 & 0.7426 & 0.7456 & 0.473 \\
 & Karla Tamil Upright & 0.7466 & 0.7162 & 0.7803 & 0.7833 & 0.542 \\
 & Meera Inimai & 0.7988 & 0.7461 & 0.7563 & 0.7594 & 0.982 \\
Bengali & Mina Bold & 0.7637 & 0.7326 & 0.7830 & 0.7690 & 0.167 \\
 & Mina & 0.7717 & 0.7472 & 0.8062 & 0.8069 & 0.177 \\
 & Tiro Bangla Italic & 0.7677 & 0.7344 & 0.7585 & 0.7592 & 0.115 \\
 & Tiro Bangla & 0.7718 & 0.7502 & 0.7762 & 0.7772 & 0.103 \\
Telugu & Chathura Bold & 0.7697 & 0.7380 & 0.7716 & 0.7893 & 0.254 \\
 & Chathura ExtraBold & 0.7788 & 0.7588 & 0.7738 & 0.7874 & 0.278 \\
 & Chathura Light & 0.7902 & 0.7653 & 0.7734 & 0.8024 & 0.155 \\
 & Chathura & 0.7872 & 0.7625 & 0.7709 & 0.7943 & 0.193 \\
 & Chathura Thin & 0.7940 & 0.7542 & 0.7795 & 0.8189 & 0.163 \\
 & Ravi Prakash & 0.7377 & 0.7021 & 0.7838 & 0.7802 & 0.122 \\
 & Sirivennela & 0.7515 & 0.7155 & 0.7778 & 0.7938 & 0.145 \\
 & Tenali Ramakrishna & 0.7436 & 0.6961 & 0.8069 & 0.8113 & 0.390 \\
Kannada & Akaya Kanadaka & 0.7127 & 0.6666 & 0.7130 & 0.7110 & 0.133 \\
 & Noto Serif Kannada & 0.6848 & 0.6303 & 0.7705 & 0.7719 & 0.143 \\
 & Noto Serif Kannada wght\,100 & 0.7442 & 0.7014 & 0.7467 & 0.7700 & 0.486 \\
 & Noto Serif Kannada wght\,200 & 0.7202 & 0.6777 & 0.7591 & 0.7736 & 0.332 \\
 & Noto Serif Kannada wght\,300 & 0.6952 & 0.6467 & 0.7711 & 0.7783 & 0.392 \\
 & Noto Serif Kannada wght\,400 & 0.6692 & 0.6231 & 0.7544 & 0.7563 & 0.209 \\
 & Noto Serif Kannada wght\,500 & 0.6585 & 0.6164 & 0.7601 & 0.7600 & 0.204 \\
 & Noto Serif Kannada wght\,600 & 0.6679 & 0.6229 & 0.7642 & 0.7603 & 0.204 \\
 & Noto Serif Kannada wght\,700 & 0.6645 & 0.6040 & 0.7547 & 0.7449 & 0.228 \\
 & Noto Serif Kannada wght\,800 & 0.6691 & 0.6287 & 0.7544 & 0.7414 & 0.303 \\
 & Noto Serif Kannada wght\,900 & 0.6595 & 0.6082 & 0.7521 & 0.7321 & 0.199 \\
\bottomrule
\end{tabular}
\end{table}

\begin{table}[p]
\centering
\small
\caption{Per-face gate scores, continued: the four grafted scripts
(Malayalam/Gujarati on gen-v3.8b; Gurmukhi/Odia on gen-v3.11b).}
\label{tab:perfaceb}
\begin{tabular}{llccccc}
\toprule
Script & Held-out face & SSIM & SSIM p25 & Tpl.\ SSIM & Morph & Style ratio \\
\midrule
Malayalam & Lohit Malayalam & 0.6996 & 0.6384 & 0.7304 & 0.7245 & 0.220 \\
 & RIT Panmana & 0.7657 & 0.7244 & 0.8150 & 0.8192 & 0.422 \\
 & RIT TN Joy Bold & 0.7482 & 0.6912 & 0.8006 & 0.8011 & 0.410 \\
 & RIT TN Joy ExtraBold & 0.7042 & 0.6500 & 0.7959 & 0.7939 & 0.277 \\
 & RIT TN Joy & 0.7332 & 0.6997 & 0.7865 & 0.7892 & 0.424 \\
 & RIT Uroob Italic & 0.7635 & 0.7154 & 0.7700 & 0.7718 & 0.087 \\
 & RIT Uroob & 0.7663 & 0.7108 & 0.7725 & 0.7743 & 0.129 \\
Gujarati & Kumar One Outline & 0.6820 & 0.6332 & 0.7068 & 0.7038 & 0.139 \\
 & Mogra & 0.7218 & 0.6792 & 0.7453 & 0.7361 & 0.154 \\
Gurmukhi & Braah One & 0.7536 & 0.7199 & 0.7618 & 0.7676 & 0.194 \\
 & Tiro Gurmukhi Italic & 0.7920 & 0.7709 & 0.8173 & 0.8176 & 0.118 \\
 & Tiro Gurmukhi & 0.8126 & 0.8022 & 0.8333 & 0.8380 & 0.170 \\
Odia & Lohit Oriya & 0.7290 & 0.6734 & 0.7619 & 0.7628 & 0.243 \\
 & Noto Serif Oriya & 0.7861 & 0.7468 & 0.8058 & 0.8091 & 0.175 \\
 & Noto Serif Oriya wght\,400 & 0.7602 & 0.7092 & 0.7893 & 0.7928 & 0.184 \\
 & Noto Serif Oriya wght\,500 & 0.7621 & 0.7087 & 0.7826 & 0.7834 & 0.169 \\
 & Noto Serif Oriya wght\,600 & 0.7758 & 0.7438 & 0.7937 & 0.7910 & 0.164 \\
 & Noto Serif Oriya wght\,700 & 0.7601 & 0.6888 & 0.7871 & 0.7754 & 0.151 \\
\bottomrule
\end{tabular}
\end{table}

\section{Preset catalogue}
\label{app:presets}
The serving deployment exposes 57 curated presets across the nine
scripts, each a fully assembled TTF generated from a named donor (Latin
display staples such as Lobster, Anton, Caveat, Bangers, Pacifico, plus
native-script donors such as BhuTuka Expanded One for Gurmukhi and
decorative Bengali/Odia families). Every preset ships with its critic
report; fleet-wide shaped-cluster accuracy is 0.981--1.000. Per-script
galleries render each preset over sample words in the target script.


\begingroup\small
\begin{thebibliography}{10}

\bibitem{googlefonts2026}
{Google Fonts} family metadata (script-subset filter).
\newblock \url{https://fonts.google.com/?subset=oriya}, family metadata
  endpoint \url{https://fonts.google.com/metadata/fonts}; accessed 3 September
  2026: 1,946 families, 6 with the Odia (\texttt{oriya}) subset, 1,825 with
  Latin, 2026.

\bibitem{harfbuzz}
Harfbuzz text shaping engine.
\newblock \url{https://harfbuzz.github.io}, 2024.

\bibitem{cfg2022}
Jonathan Ho and Tim Salimans.
\newblock Classifier-free diffusion guidance.
\newblock In {\em NeurIPS Workshop}, 2021.

\bibitem{lora2022}
Edward~J Hu, Yelong Shen, Phillip Wallis, Zeyuan Allen-Zhu, Yuanzhi Li, Shean
  Wang, and Weizhu Chen.
\newblock Lora: Low-rank adaptation of large language models.
\newblock In {\em ICLR}, 2022.

\bibitem{vecglypher2026}
Xiaoke Huang, Bhavul Gauri, Kam~Woh Ng, Tony Ng, Mengmeng Xu, Zhiheng Liu,
  Weiming Ren, Zhaochong An, Zijian Zhou, Haonan Qiu, Yuyin Zhou, Sen He,
  Ziheng Wang, Tao Xiang, and Xiao Han.
\newblock Vecglypher: Unified vector glyph generation with language models.
\newblock {\em arXiv preprint arXiv:2602.21461}, 2026.

\bibitem{funit2019}
Ming-Yu Liu, Xun Huang, Arun Mallya, Tero Karras, Timo Aila, Jaakko Lehtinen,
  and Jan Kautz.
\newblock Few-shot unsupervised image-to-image translation.
\newblock In {\em ICCV}, 2019.

\bibitem{dualvector2023}
Ying-Tian Liu, Zhifei Zhang, Yuan-Chen Guo, Matthew Fisher, Zhaowen Wang, and
  Song-Hai Zhang.
\newblock Dualvector: Unsupervised vector font synthesis with dual-part
  representation.
\newblock In {\em CVPR}, 2023.

\bibitem{t2iadapter2023}
Chong Mou, Xintao Wang, Liangbin Xie, Yanze Wu, Jian Zhang, Zhongang Qi, Ying
  Shan, and Xiaohu Qie.
\newblock T2i-adapter: Learning adapters to dig out more controllable ability
  for text-to-image diffusion models.
\newblock In {\em AAAI}, 2024.

\bibitem{odonovan2014}
Peter O'Donovan, J{\=a}nis L{\=\i}beks, Aseem Agarwala, and Aaron Hertzmann.
\newblock Exploratory font selection using crowdsourced attributes.
\newblock {\em ACM TOG (SIGGRAPH)}, 2014.

\bibitem{census2011lang}
{Office of the Registrar General \& Census Commissioner, India}.
\newblock {Census of India 2011, Paper 1 of 2018: Language --- India, States
  and Union Territories (Table C-16)}.
\newblock \url{https://censusindia.gov.in/nada/index.php/catalog/42458}, 2018.

\bibitem{mxfont2021}
Song Park, Sanghyuk Chun, Junbum Cha, Bado Lee, and Hyunjung Shim.
\newblock Multiple heads are better than one: Few-shot font generation with
  multiple localized experts.
\newblock In {\em ICCV}, 2021.

\bibitem{qwen25vl}
{Qwen Team}.
\newblock Qwen2.5-{VL} technical report.
\newblock {\em arXiv:2502.13923}, 2025.

\bibitem{ldm2022}
Robin Rombach, Andreas Blattmann, Dominik Lorenz, Patrick Esser, and Bj{\"o}rn
  Ommer.
\newblock High-resolution image synthesis with latent diffusion models.
\newblock In {\em CVPR}, 2022.

\bibitem{fontclip2024}
Yuki Tatsukawa, I-Chao Shen, Anran Qi, Yuki Koyama, Takeo Igarashi, and Ariel
  Shamir.
\newblock Fontclip: A semantic typography visual-language model for
  multilingual font applications.
\newblock In {\em Eurographics}, 2024.

\bibitem{vecfusion2024}
Vikas Thamizharasan, Difan Liu, Shantanu Agarwal, Matthew Fisher, Micha{\"e}l
  Gharbi, Oliver Wang, Alec Jacobson, and Evangelos Kalogerakis.
\newblock Vecfusion: Vector font generation with diffusion.
\newblock In {\em CVPR}, 2024.

\bibitem{unicode16}
{Unicode Consortium}.
\newblock The {Unicode} standard, version 16.0 --- chapter 12: South and
  central asia-{I}.
\newblock \url{https://www.unicode.org/versions/Unicode16.0.0/}, 2024.

\bibitem{cffont2023}
Chi Wang, Min Zhou, Tiezheng Ge, Yuning Jiang, Hujun Bao, and Weiwei Xu.
\newblock Cf-font: Content fusion for few-shot font generation.
\newblock In {\em CVPR}, 2023.

\bibitem{vecfontsdf2023}
Zeqing Xia, Bojun Xiong, and Zhouhui Lian.
\newblock Vecfontsdf: Learning to reconstruct and synthesize high-quality
  vector fonts via signed distance functions.
\newblock In {\em CVPR}, 2023.

\bibitem{dgfont2021}
Yangchen Xie, Xinyuan Chen, Li~Sun, and Yue Lu.
\newblock Dg-font: Deformable generative networks for unsupervised font
  generation.
\newblock In {\em CVPR}, 2021.

\bibitem{fontdiffuser2024}
Zhenhua Yang, Dezhi Peng, Yuxin Kong, Yuyi Zhang, Cong Yao, and Lianwen Jin.
\newblock Fontdiffuser: One-shot font generation via denoising diffusion with
  multi-scale content aggregation and style contrastive learning.
\newblock In {\em AAAI}, 2024.

\bibitem{controlnet2023}
Lvmin Zhang, Anyi Rao, and Maneesh Agrawala.
\newblock Adding conditional control to text-to-image diffusion models.
\newblock In {\em ICCV}, 2023.

\end{thebibliography}
\endgroup
\end{document}